%% file: main.tex
\documentclass[10pt,twocolumn,letterpaper]{article}

\usepackage{cvpr}              

\input{preamble}

\definecolor{cvprblue}{rgb}{0.21,0.49,0.74}
\usepackage[pagebackref,breaklinks,colorlinks,allcolors=cvprblue]{hyperref}
\usepackage{svg}
\usepackage{dblfloatfix}
\usepackage{caption}
\usepackage{mathrsfs}
\svgpath{{figs/}}
\newcommand{\modelname}{FoundHand\xspace}
\newcommand{\datasetname}{FoundHand-10M\xspace}

\def\paperID{2890} 
\def\confName{CVPR}
\def\confYear{2025}

\title{\modelname: Large-Scale Domain-Specific Learning for\\ Controllable Hand Image Generation}

\author{Kefan Chen\textsuperscript{*}\textsuperscript{1}\textsuperscript{2}
\and
Chaerin Min\textsuperscript{*}\textsuperscript{1}
\and
Linguang Zhang\textsuperscript{2}
\and
Shreyas Hampali\textsuperscript{2}
\and
Cem Keskin\textsuperscript{2}
\and
Srinath Sridhar\textsuperscript{1}
\\ 
\centerline{%
\textsuperscript{1}Brown University
\hspace{2em} 
\textsuperscript{2}Meta Reality Labs}
}

\begin{document}
\twocolumn[{
    \maketitle
    \includegraphics[width=\textwidth]{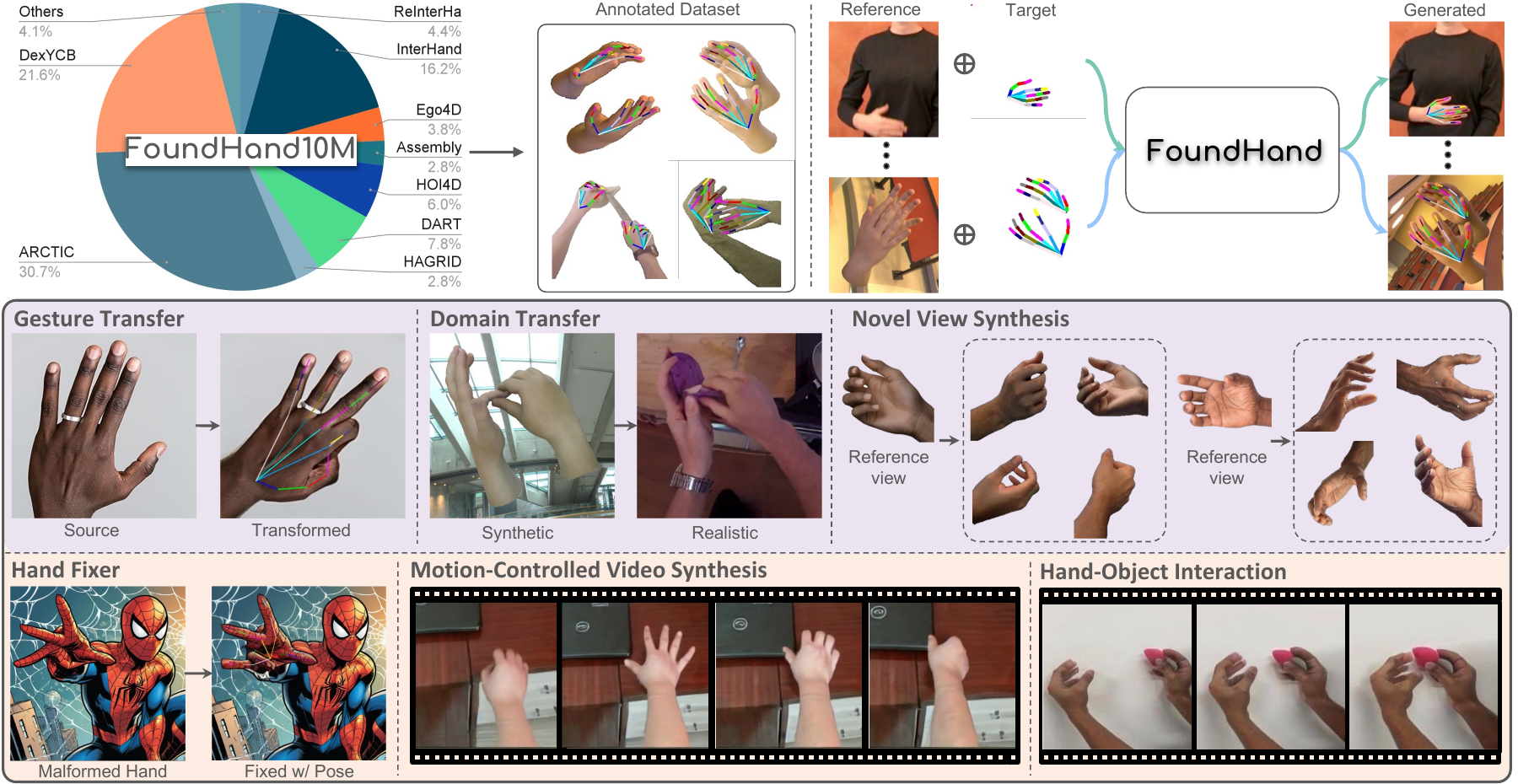}
    \captionof{figure}{We present \textbf{\modelname}, a domain-specific image generation model that can synthesize realistic single and dual hand images.
    \modelname is trained on our large-scale \textbf{\datasetname} dataset which contains automatically extracted 2D keypoints and segmentation mask annotations (top left).
    \modelname is formulated as a 2D pose-conditioned image-to-image diffusion model that enables precise hand pose and camera viewpoint control (top right).
    Optionally, we can condition the generation with a reference image to preserve its style (top right).
    Our model demonstrates exceptional in-the-wild generalization across hand-centric applications and has  \textcolor{Orchid}{core capabilities} such as gesture transfer, domain transfer, and novel view synthesis (middle row).
    This endows \modelname with \textcolor{Apricot}{zero-shot applications} to fix malformed hand images and synthesize coherent hand and hand-object videos, without explicitly giving object cues (bottom row).
    }
    \label{fig:teaser}
\vspace{0.2in}
}]
\footnotetext[1]{\textsuperscript{*}Equal contribution.}
\footnotetext[2]{Project Page: \url{https://ivl.cs.brown.edu/research/foundhand.html}}
\input{sec/0_abstract}  
\input{sec/1_intro}

\input{sec/2_related_work}
\input{sec/3_method}
\input{sec/4_experiment}

\input{sec/5_applications}

\input{sec/6_conclusion}
{
    \small
    \bibliographystyle{ieeenat_fullname}
    \bibliography{main}
}

\end{document}

%% file: preamble.tex
\newcommand{\srinath}[1]{}
\newcommand{\linguang}[1]{}
\newcommand{\arthur}[1]{}
\newcommand{\chaerin}[1]{}

\usepackage{multirow}
\usepackage{array}
\usepackage{tabularx}
\usepackage[dvipsnames]{xcolor}
\usepackage{tikz}
\definecolor{PURP}{HTML}{782A78}
\definecolor{ORAN}{HTML}{FBCC96}

%% file: sec/0_abstract.tex
\vspace{0.1in}
\begin{abstract}
Despite remarkable progress in image generation models, generating realistic hands remains a persistent challenge due to their complex articulation, varying viewpoints, and frequent occlusions.
We present \modelname, a large-scale domain-specific diffusion model for synthesizing single and dual hand images.
To train our model, we introduce \datasetname, a large-scale hand dataset with 2D keypoints and segmentation mask annotations.
Our insight is to use 2D hand keypoints as a universal representation that encodes both hand articulation and camera viewpoint.
\modelname learns from image pairs to capture physically plausible hand articulations, natively enables precise control through 2D keypoints, and supports appearance control.
Our model exhibits core capabilities that include the ability to repose hands, transfer hand appearance, and even synthesize novel views.
This leads to zero-shot capabilities for fixing malformed hands in previously generated images, or synthesizing hand video sequences.
We present extensive experiments and evaluations that demonstrate state-of-the-art performance of our method.


%
\end{abstract}

%% file: sec/1_intro.tex
\section{Introduction}
\label{sec:intro}
Recent years have witnessed remarkable progress in generative models~\cite{NIPS2014_5ca3e9b1,Karras2018ASG}, particularly diffusion models~\cite{DDPM,ddim,Peebles2022DiT,improvedDDPM} that can generate photorealistic images of scenes~\cite{animate_anyone,sajnani2024geodiffuser,rombach2021highresolution,diffusionbeatsgan,dalle,dalle2}.
These models have also been used to generate human faces~\cite{chan2022efficient,Huang2020LearningIM, Nirkin2019FSGANSA,Deng2020DisentangledAC,Ren2021PIRendererCP}, or bodies~\cite{ma2017pose, Chen_2019_CVPR_Workshops,total_gen, Siarohin2017DeformableGF, Wang2019ExampleGuidedSI,Albahar2021PoseWS,Karras2018ASG, 3d_cloth, 4drecons}.
However, popular generative models~\cite{rombach2021highresolution,ramesh2021zero,Midjourn99:online} consistently struggle to generate realistic images of hands~\cite{WhyAIart15:online,WhyAIcan9:online} -- a critical shortcoming given the need for hand images in art, interaction~\cite{manus}, mixed reality~\cite{hand_vr}, and robotics~\cite{constrained_grasp, vihope, dexgraspnet, cgdf}.

This shortcoming can be attributed to the highly articulated nature of hand motion, and the complexity of hand-object interactions.
Large-scale image datasets~\cite{schuhmann2022laion} contain a wide range of object classes.
As a result, hands are not well-sampled in these datasets -- they either occupy very few pixels, or complex articulations are not captured.
Thus, generative model performance on hands~\cite{controlnet,controlnext} is poor (see \Cref{sec:experiment}).
Furthermore, current models lack representations that can reliably control the generation of complex hand articulations.
Recent methods have attempted to use intermediate 3D hand representations~\cite{mano}, but these are not always available, or are inaccurate~\cite{narasimhaswamy2024handiffuser,Hu2021ModelAwareGT,lu2023handrefiner,attentionhand,handcraft}. 

In this paper, we address these issues by introducing an image generation model that can synthesize realistic hands in a controllable manner.
Our first insight to improve performance on hands is to build a \emph{domain-specific model} with large-scale~\cite{singh2023effectiveness,oquab2023dinov2,autoregressive-image-models,sapiens} hand data.
Since no such dataset exists, we build \textbf{\datasetname}, a large-scale hand dataset created from existing video and multi-view hand datasets~\cite{ego4d,epickitchen,interhand,hoi4d,reinterhand,wlasl,dart,renderih,hagrid,zb2017hand,assemblyhands,dexycb,arctic}.
%
Our second insight is that \emph{2D keypoints} are an easily-accessible representation of hand pose that can be accurately obtained at scale.
Since 2D keypoints are projections of 3D hand joints onto a camera, they naturally encode both articulation and camera viewpoint, making them a suitable representation for controllable generation.
Therefore, our high-quality dataset contains a diverse collection of 10M images with automatically extracted 2D keypoint and segmentation masks annotations~\cite{kirillov2023segment, mediapipe}.



With the help of our large-scale dataset, we then build \textbf{\modelname}, a diffusion model that can generate single or dual hand images precisely controlled using 2D keypoint heatmap conditioning.
Optionally, our model can also take a reference image as an additional condition to preserve output appearance.
Inspired by video diffusion models~\cite{moviegen, dimensionx, sparsectrl, cogvideox},
we formulate generation as an image-to-image translation task that achieves the benefits of video models without the computational burden.
We achieve this by training the model on pairs of frames from hand videos or multi-view images to learn pose and viewpoint transformation.
\modelname builds upon a latent vision transformer backbone~\cite{Peebles2022DiT,rombach2021highresolution} pre-trained on ImageNet~\cite{imagenet} for robust visual feature generalization.
We employ 3D self-attention between input image pairs, which, while computationally intensive for video models, remains tractable in our two-frame setting.
Our model learns hand feature embeddings by encoding spatially-aligned input modalities: image latent, 2D keypoint heatmaps, and hand segmentation masks.
This multi-modal alignment grounds the model in hand physicality, enhancing its structural understanding and ensuring realistic generation that adheres to pose conditions.

Our trained \modelname model has multiple core capabilities that enable versatile applications through simple 2D keypoint control.
Users can \textbf{repose hands} in images while retaining the appearance of the original image, \textbf{transfer hand appearance} from one style to another, and even \textbf{synthesize novel views} without explicit 3D supervision.
To our surprise, \modelname achieves remarkable generalization across novel poses and appearance, extending even to diverse artistic styles and hands with accessories.
Our model also demonstrates \textbf{zero-shot capabilities} in (1)~fixing malformed hands generated using previous models, and (2)~synthesizing motion-controlled videos without training on video sequences, (3)~generating hand-object interaction (HOI) videos without explicit training on any objects, indicating a deep understanding of hand structure and dynamics.
We evaluate our model through extensive experiments on in-the-wild test data and AI generated images, demonstrating state-of-the-art performance.
To sum up:
\begin{itemize}
\item We introduce \textbf{\datasetname}, a large-scale hand dataset containing over 10M hand images with annotated 2D keypoints and segmentation masks.
%
\item \textbf{\modelname}, a generative model that can synthesize realistic single or dual hand images.
This model is natively controlled using 2D keypoint representation enabling precise articulation and camera viewpoint control.
%
\item A variety of \textbf{core capabilities} including hand reposing, domain transfer, and 3D novel view synthesis.
These capabilities enable zero-shot applications including fixing malformed hands, video generation, and HOI video generation with deformable objects. 
%
%
\end{itemize}






%% file: sec/2_related_work.tex
\section{Related Works}
\label{sec:related_work}

\paragraph{Generative Models.}
Diffusion models~\cite{DDPM,improvedDDPM,ddim} have gained immense advancement on general text-to-image generation~\cite{stablediffusion, imagen}, powered by large language models~\cite{bert, clip}. 
In response, spatially-conditioned image generation models~\cite{controlnet,sun2024anycontrol} were proposed.
Closer to our method, some previous works suggested to build generative models specifically for faces~\cite{Huang2020LearningIM,Nirkin2019FSGANSA, 3cfaceCAM, towards_3d_face, Deng2020DisentangledAC,Ren2021PIRendererCP} and bodies~\cite{ma2017pose, Chen_2019_CVPR_Workshops,total_gen, Siarohin2017DeformableGF, Wang2019ExampleGuidedSI}.
Only a few such models exist for hands~\cite{handiffuser, hand1000,attentionhand}, but they are not trained at scale since they rely on 3D hand models~\cite{mano} that may not always be available or accurate.
Other methods~\cite{coshand,handcraft,lu2023handrefiner,affordance_diffusion,graspdiffusion,animate_anyone} learn to generate hand images of interaction, correction, or video.
However, these methods lack versatility and can only work with each specific task.
\paragraph{Datasets for Hands.}
While datasets with hands are relatively sparse than general-purpose image datasets~\cite{imagenet,schuhmann2022laion,lin2014microsoft}, there are a few datasets that contains hands.
Some datasets~\cite{ego4d, wlasl, hagrid} proposed videos of hands, while others~\cite{interhand,reinterhand,arctic,dexycb,diva360} released multi-view datasets of hands.
Egocentric hand dataset is proposed by ~\cite{ego4d, hoi4d, epickitchen}, and exocentric hand images are provided by ~\cite{wlasl, hagrid, rhd}.
Classified realism, ~\cite{rhd,renderih,dart} includes synthetic data, so they provide accurate 3D annotation.
On the other hand, real-world datasets ~\cite{hagrid,wlasl,assemblyhands, ego4d, hoi4d, arctic, dexycb} have their own types of annotations, including mesh, 3D keypoints, 2D keypoints, and depth.
Despite multiple different datasets, there has yet been an integrated large dataset that incorporate all this diversity with a single type and convention of annotations.

\begin{figure*}[!ht]
    \centering
    \includegraphics[width=\textwidth]{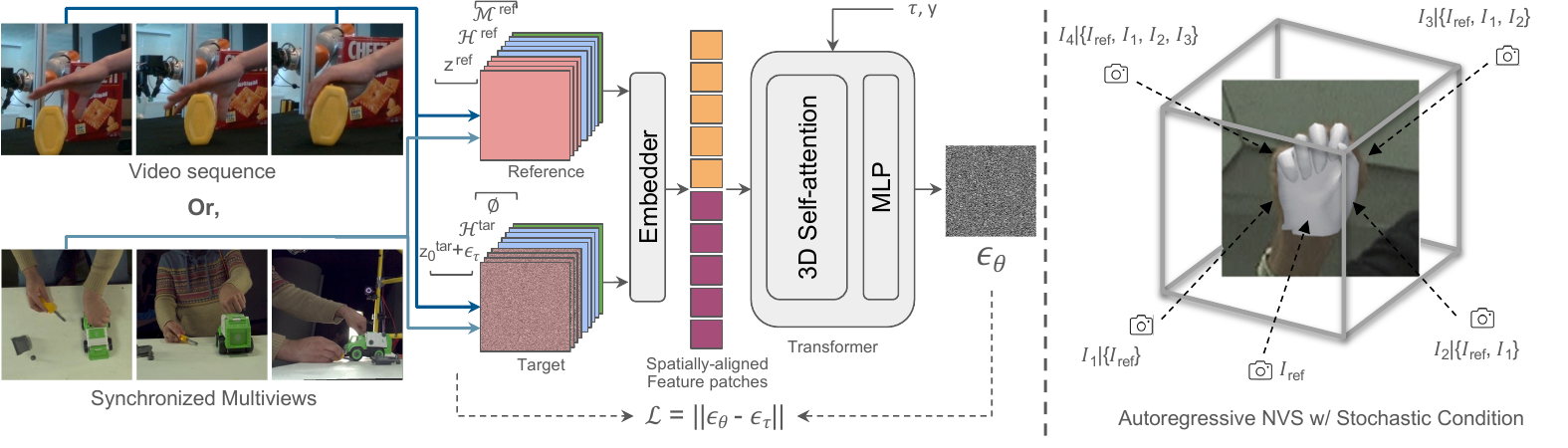}
    \caption{(Left)~During training, we randomly sample two frames from a video sequence or two different views of a frame as the reference and target frame and encode them using a pretrained VAE as the latent diffusion model. We concatenate the encoded image features $z$ with keypoint heatmaps $\mathcal{H}$ and hand mask $\mathcal{M}$ and encode them with a shared-weight embedder to acquire spatially-aligned feature patches before feeding to transformer with 3D self-attention. The target hand mask is set to $\emptyset$ since it is not required at test time.
    $y$ indicates if the two frames are from the synchronized views.
    } 
    \label{fig:method}
\end{figure*}
\vspace{-1.5em}
\paragraph{Generative Models for Hands.}
Despite important applications in human interaction understanding, mixed reality, and robotics, 
only a few works~\cite{handiffuser, attentionhand, genheld} attempted to generate hand images with background, but they require 3D MANO~\cite{mano} hands as condition.
Others~\cite{signllm, dreamoving} specialize in sign languages and dance, respectively.
Early work~\cite{Tang2018GestureGANFH} introduced gesture transfer, with poor generalization ability in modern standard.
Models with spatial controls~\cite{controlnet,sun2024anycontrol, qin2023unicontrol} provide hand image generation model trained to be conditioned on OpenPose~\cite{openpose} image.
Recently, ~\cite{coshand} proposes to transfer hand poses with deformable objects, given target hand masks.
However, hand mask and 2D keypoint images can be ambiguous where hand has heavy occlusion between fingers, while our proposed method proposes multi-channel heatmaps of 2D keypoints to disentangle each keypoint.
Recent works~\cite{lu2023handrefiner,realishuman, handcraft} tried synthesizing correct hands given malformed hands, but they rely on the off-the-shelf hand reconstruction models~\cite{meshgraphormer, dwpose,mediapipe, hamer} that less robust to malformed hands.
For video generation, ~\cite{controlnext,animate_anyone,dreamoving} provide video generation specifically trained on Openpose~\cite{openpose} images as conditions, while being limited to only two-hand generations.
To conclude, although hand generation methods specific on each different tasks have been of interests, a more versatile and robust hand generative model at scale is yet to be developed.

%% file: sec/3_method.tex
\section{Background}
\label{sec:background}
We provide a brief overview of diffusion models~\cite{Peebles2022DiT,DDPM,ho2022classifierfree} that we build upon.
Denoising diffusion probabilistic models (DDPM) have emerged as a powerful framework for generative modeling~\cite{DDPM,song2021scorebased,sohldickstein2015deep}.
The diffusion process consists of two phases: a forward process that gradually adds Gaussian noise to real data $z_0$, defined as $q(z_\tau|z_0) = \mathcal{N}\big(\sqrt{\bar{\alpha}_\tau} z_0, (1-\bar{\alpha}_\tau)I\big)$ where $\bar{\alpha}_\tau$ are predefined noise schedules, and a learned reverse process that progressively denoises corrupted samples. This reverse process $p_\theta(z_{\tau-1}|z_\tau)=\mathcal{N}\big(\mu_\theta(z_\tau), \Sigma_\theta(z_\tau)\big)$ is parameterized by a neural network $\epsilon_\theta$ trained to minimize the objective $\mathcal{L} = \mathbb{E}_{\epsilon_\tau, \tau} \left[ \parallel \epsilon_\theta(z_\tau;\tau, c) - \epsilon_\tau \parallel^2_2 \right]$, where $c$ represents conditioning information. This framework has achieved remarkable success in conditional generation tasks, particularly in text-to-image models~\cite{dalle2,clip,stablediffusion}. To enhance conditioning fidelity while maintaining sample diversity, classifier-free guidance (CFG)~\cite{ho2022classifierfree} is commonly employed using $\hat{\epsilon_\theta}=w\epsilon_\theta(z_\tau;\tau, c)+(1-w)\epsilon_\theta(z_\tau;\tau,\emptyset)$, where $w$ is a guidance scale. For computational efficiency, images are encoded in a latent space $z_0=\mathcal{E}(\mathcal{I})$ using a pre-trained variational autoencoder $\mathcal{E}$ (VAE)~\cite{rombach2021highresolution}.

\section{Domain-Specific Hand Image Generation}
\label{sec:method}
Our goal is to build a large-scale generative model for generalizable and versatile hand image generation.
To achieve this, we build three components: a large-scale hand dataset (\Cref{sec:dataset}), a pose-conditioned diffusion model (\Cref{sec:model}), and core capabilities that enable numerous downstream applications (\Cref{sec:core}).

\subsection{\datasetname}
\label{sec:dataset}
To benefit from scaling up a generative model for hands, we need a training dataset that is sufficiently large.
Existing datasets are too small, or contain similar looking images that impacts generalization.
Therefore, we build \textbf{\datasetname} by combining the RGB images from multiple existing hand datasets including DexYCB~\cite{dexycb}, ARCTIC~\cite{arctic}, ReInterHand~\cite{reinterhand}, InterHand~\cite{interhand}, Ego4D~\cite{ego4d}, AssemblyHand~\cite{ohkawa:cvpr23}, HOI4D~\cite{hoi4d}, RHD~\cite{rhd}, RenderIH~\cite{renderih}, DART~\cite{dart}, HAGRID~\cite{hagrid}, and WLASL~\cite{wlasl}.
These datasets collectively capture bimanual hand-object manipulation, hand-hand interactions, daily activities, and sign language gestures, across laboratory-controlled environments and in-the-wild scenarios through multi-view, egocentric, and third-person perspectives.

While combining RGB images is straightforward, establishing a common representation for hand pose across datasets is challenging.
One option is to use 3D representations like MANO~\cite{mano}, but it can be computationally expensive to acquire at scale and often suffer from misalignment.
In contrast, 2D keypoints provide an efficient and reliable representation that naturally encodes both articulation and viewpoint information.
Therefore, we obtain high-quality 2D keypoint and segmentation annotations using MediaPipe~\cite{mediapipe} and SAM~\cite{kirillov2023segment} for all 10M images in our dataset.
Our dataset captures diverse poses, viewpoints, lighting, and hand accessories so that our model trained on \datasetname demonstrates exceptional generalization to in-the-wild scenarios and novel appearances.
Please see \Cref{fig:teaser} (top left) and supplemental for more details.

\subsection{\modelname}
\label{sec:model}
\paragraph{Hand Representation.}
%
Different from existing generative models, we want to build a hand-specific generative model that natively supports pose and camera control.
We adopt 2D keypoints as the universal representation for hand, providing precise articulation control while enabling integration across diverse hand-related tasks. 
Rather than using a vector of 2D keypoints $\mathbf{p}\in \mathbb{R}^{42\times2}$, we encode them as Gaussian heatmaps~\cite{tompson2014real} $\mathcal{H}={\mathcal{N}(\mathbf{x}|\mathbf{p},\sigma^2)}$, where the first 21 channels $\mathcal{H}_{[0:21]}$ correspond to the right hand and the last 21 channels $\mathcal{H}_{[21:42]}$ to the left hand, with zero-valued heatmaps indicating absent hands. We then spatially align multiple input modalities: image features extracted by the pretrained VAE from StableDiffusion~\cite{stablediffusion}, keypoint heatmaps, and hand segmentation masks.
This multi-modal alignment provides crucial signals for the model to effectively learn the intricate relationships between hand appearance, silhouette, and articulation.
To facilitate subsequent model training, these aligned modalities are processed with a shared embedder that learns spatially-coherent features.

\paragraph{Model Architecture.}
The \modelname architecture (\Cref{fig:method}) builds upon a latent diffusion vision transformer backbone (DiT)~\cite{Peebles2022DiT,rombach2021highresolution}.
While the embedding layer is initialized from scratch, we leverage ImageNet-pretrained transformer blocks to benefit from rich visual priors across diverse objects and scenes, enhancing model generalization. 
Our approach models hand image generation as an image-to-image translation task, learning physically plausible hand transformations while preserving both hand appearance and scene context. 
The model processes two inputs: reference and target.
The reference comprises image features, keypoint heatmaps, and masks $[z^{ref}, \mathcal{H}^{ref}, \mathcal{M}^{ref}]\in\mathbb{R}^{H \times W \times C}$.
During training, we add noise to the target image features by scheduled noise steps $\tau$, as follows, $[z_0^{tar}+\epsilon_\tau, \mathcal{H}^{tar}, \emptyset]$ where $\epsilon_\tau$ is the sampled noise.
At inference time, the image feature channels of the target are iteratively denoised from $z_\tau\sim\mathcal{N}(0, \bold{I})$, guided by the provided keypoint heatmaps.
We employ classifier-free guidance~\cite{ho2022classifierfree} to balance generation quality with control fidelity.

%
The model processes these inputs in a spatially-aligned multi-modal manner.
After passing them through a shared-weight embedder, we utilize 3D self-attention between the reference and target frames in the transformer block, enabling effective cross-frame feature alignment and simultaneous learning of both pose and view transformations. Finally, the model outputs the reconstructed noise $\epsilon_\theta$, which is then compared with $\epsilon_\tau$ using the L2 loss.
%

\paragraph{Training.}
Our training strategy
leverages both sequential and multi-view data available in \datasetname sourced from existing datasets. 
Specifically, we learn hand pose transformations by sampling frame pairs from temporal sequences, and viewpoint transformations by sampling synchronized frames from multiple camera views. 
We utilize a binary flag $y$ to differentiate the handling of pose transformation versus view transformation tasks, enabling the model to adapt its generation strategy based on the intended application.
To enhance robustness and generalization, we first apply comprehensive data augmentation, including random gamma correction, hand swapping, horizontal flipping (to let the model learn handedness transformation), and cropping. 
Second, we implement condition dropout by randomly masking either all reference tokens or target heatmaps, enabling the model to learn both conditional and marginal distributions. 
This not only enhances robust understanding of each individual condition's effect, but also improves the versatility of our model in various tasks by allowing the reference keypoint heatmaps and masks to be optional during inference.
Third, we leverage REPA alignment~\cite{yu2024repa}, which aligns intermediate features with self-supervised representations from DINOv2~\cite{oquab2023dinov2}, improving training efficiency and generation quality.



\vspace{0.1in}
\subsection{Core Generation Capabilities}
\label{sec:core}
%
Through the unified 2D keypoint conditioning, and our model architecture, we can enable versatile core generation capabilities as described below.
Please see \Cref{sec:experiment} for details on how these capabilities enable new applications.

%
\paragraph{Gesture and Domain Transfer.} 
FoundHand enables precise hand image manipulation through a flexible conditioning framework that combines reference images and target 2D keypoints. 
Given a reference hand image and target keypoints, our model synthesizes transformed hand gestures while faithfully preserving the fine-grained appearance details, background context, and target keypoints.
Beyond pose manipulation, FoundHand facilitates appearance transfer across domains, by maintaining fixed 2D keypoint conditions while varying the reference image to reflect the target domain's style. 
Although our training pipeline incorporates 2D hand masks as auxiliary supervision, the trained model can operate without mask inputs for either reference or target conditions, enhancing its practical applicability.

\paragraph{Novel View Synthesis (NVS).}
\modelname achieves 3D-consistent novel view synthesis through
learned anatomical priors, despite being trained without explicit 3D supervision.
Our key insight is that 2D keypoints naturally encode both hand articulation and camera viewpoint information, eliminating the need for explicit camera parameters during training or inference.
This represents an advantage over existing novel view synthesis~\cite{nerf,ingp,kerbl3Dgaussians} and sparse view reconstruction methods~\cite{vivid1to3,watson2022novel,genvs,reconfusion,sparsefusion,tsdf,viewneti,renderdiffusion}, which often struggle with generalization to out-of-distribution camera viewpoints.

Given a reference view, our pipeline first estimates 3D joints $\mathcal{J}\in\mathbb{R}^{42\times3}$ using an off-the-shelf 3D pose estimator~\cite{hamer} assuming a camera projection $\mathcal{K}$, then projects these 3D joints into target cameras to obtain conditioning 2D keypoints $\mathcal{K}(\mathcal{J})\in\mathbb{R}^{42\times2}$.
To enhance 3D consistency, we extend our single-image conditioning model through stochastic conditioning~\cite{watson2022novel} where at each denoising step $\tau$, the next latent state is sampled according to $z_{\tau-1} \sim p_\theta(z_{\tau-1}|z_{\tau}, \mathcal{K}(\mathcal{J}), \mathcal{E}(\mathcal{I}),y=1)$, where $\mathcal{I}$ is randomly sampled from the set $\{\mathcal{I}_{ref}, \mathcal{I}_1, ..., \mathcal{I}_{n-1}\}$ of reference and previously generated views. 
For stable generation, we initialize the process by sampling new camera views within a small cone around the reference camera $\mathcal{I}_{ref}$, then follow an autoregressive generation process (\Cref{fig:method}) to sample the subsequent views $\mathcal{I}_n$ by conditioning on both the reference and previously generated views: $\mathcal{I}_n|\{\mathcal{I}_{ref}, \mathcal{I}_1, ..., \mathcal{I}_{n-1}\}$. Our approach enables robust generalization to new viewpoints, leveraging learned priors to resolve 2D projection ambiguities.

%% file: sec/4_experiment.tex
\begin{figure*}[!tp]
    \renewcommand{\tabcolsep}{0pt}
    \renewcommand{\arraystretch}{0.}
    \centering \footnotesize
    \begin{tabular}{cccccc}
        \includegraphics[width=0.16\textwidth]{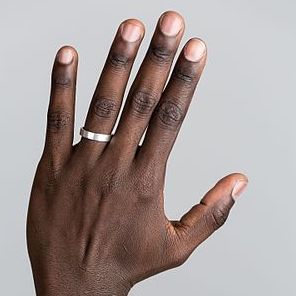}&
        \includegraphics[width=0.16\textwidth]{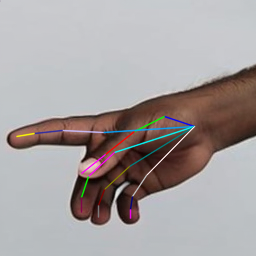}&
        \includegraphics[width=0.16\textwidth]{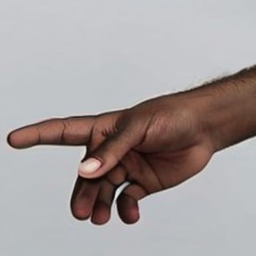}&
        \includegraphics[width=0.16\textwidth]{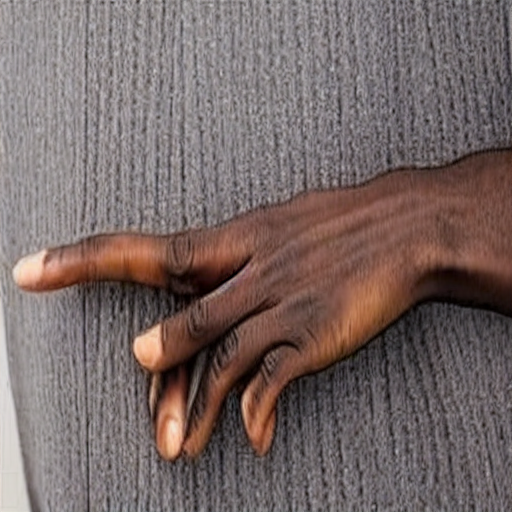} &
        \includegraphics[width=0.16\textwidth]{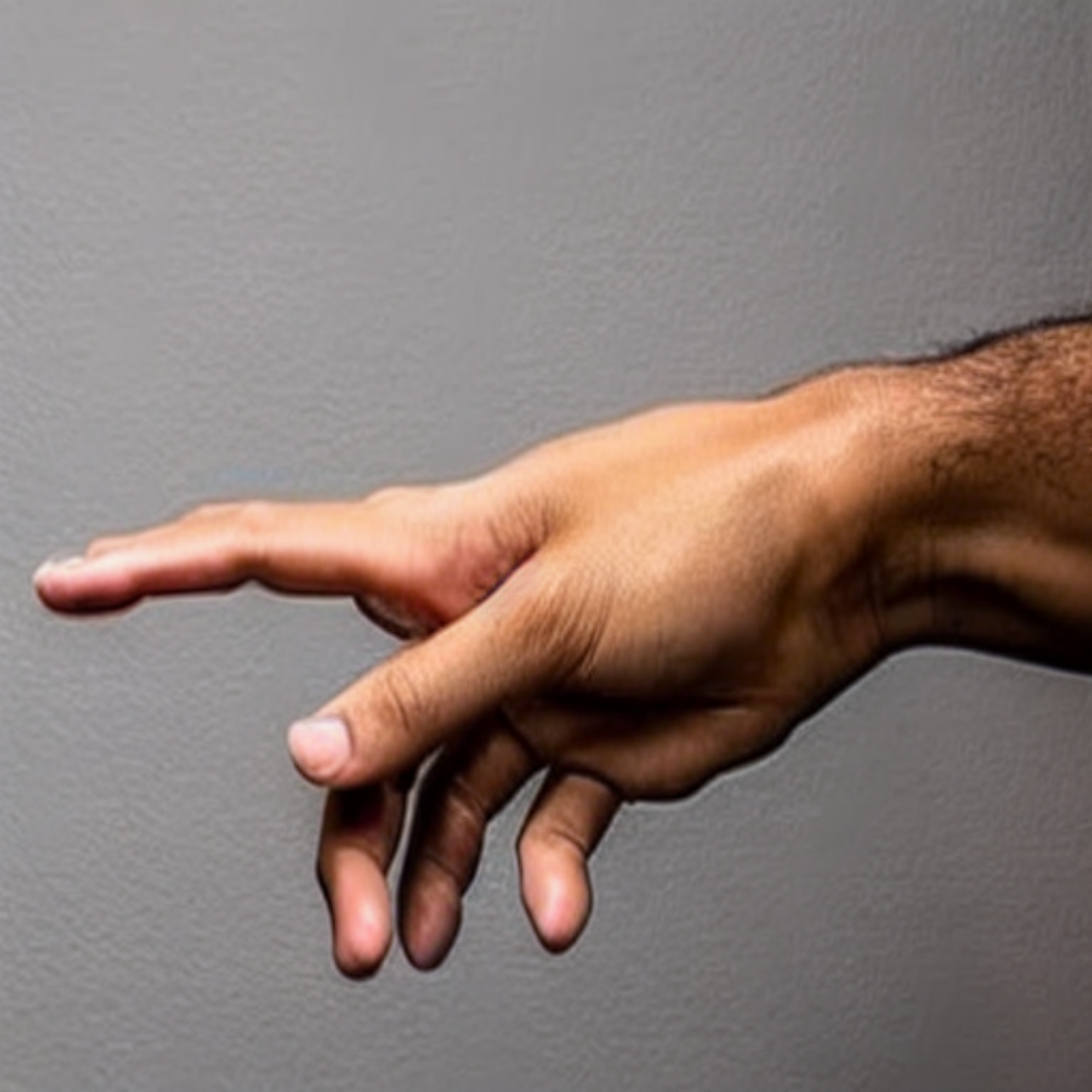}&
        \includegraphics[width=0.16\textwidth]{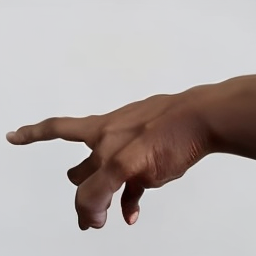} \\
        
        \includegraphics[width=0.16\textwidth]{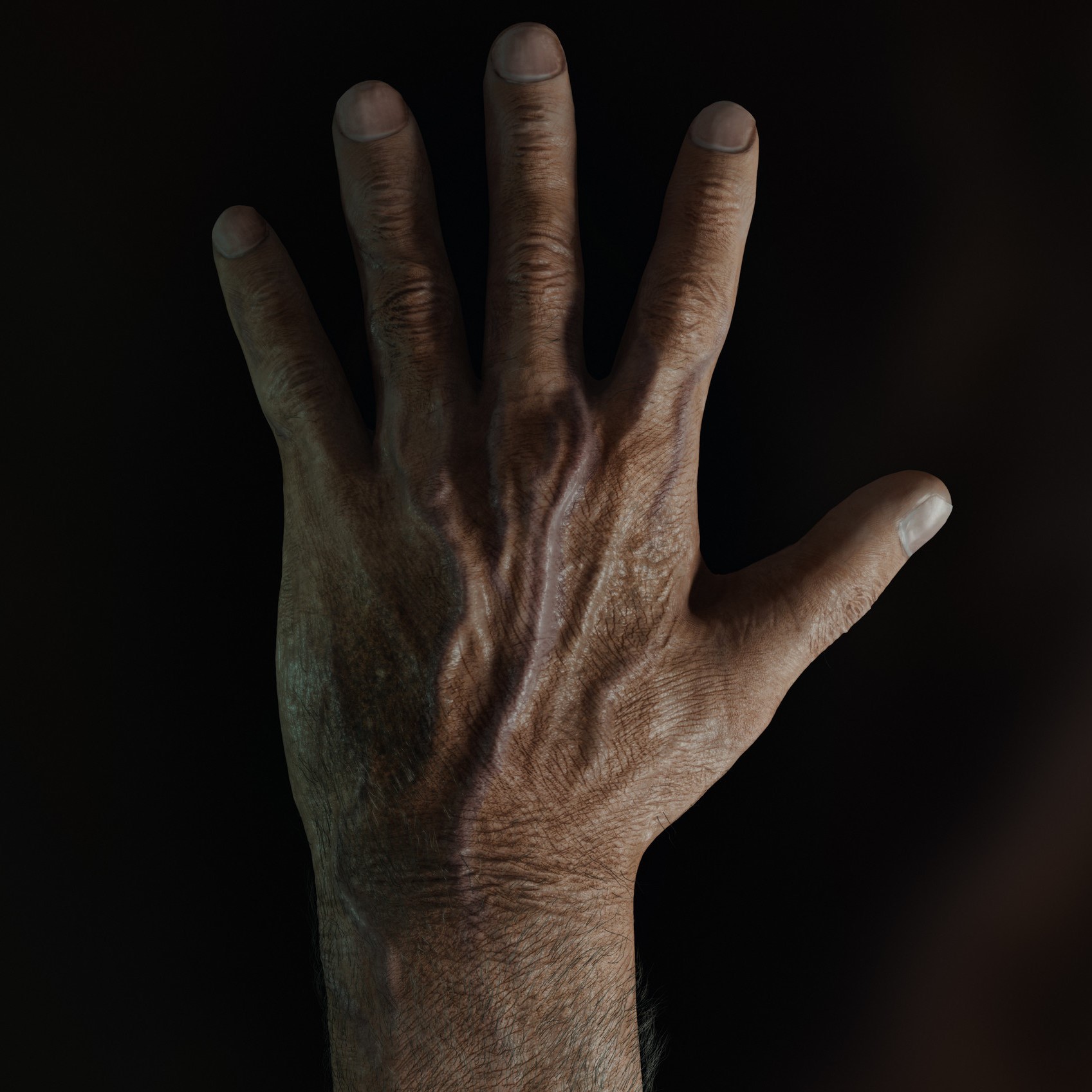}&
        \includegraphics[width=0.16\textwidth]{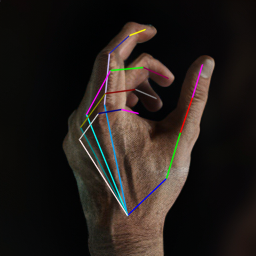}&
        \includegraphics[width=0.16\textwidth]{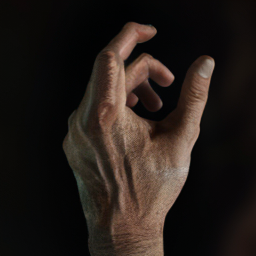}&
        \includegraphics[width=0.16\textwidth]{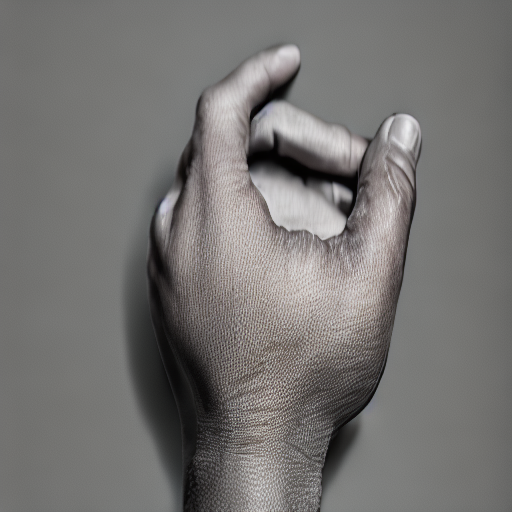} &
        \includegraphics[width=0.16\textwidth]{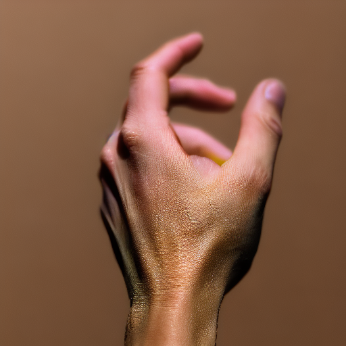} & 
        \includegraphics[width=0.16\textwidth]{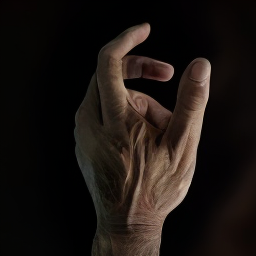}\\

        \includegraphics[width=0.16\textwidth]{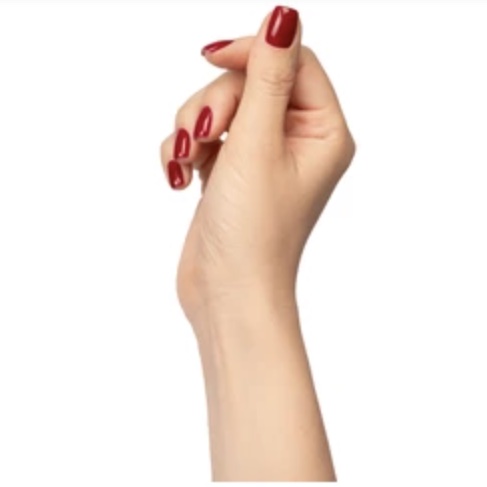}&
        \includegraphics[width=0.16\textwidth]{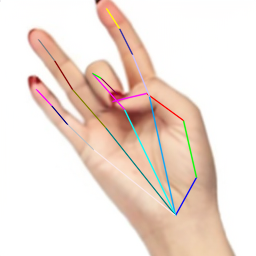}&
        \includegraphics[width=0.16\textwidth]{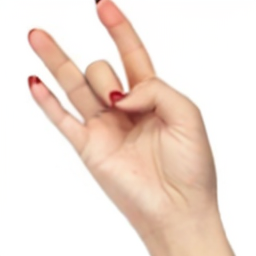}&
        \includegraphics[width=0.16\textwidth]{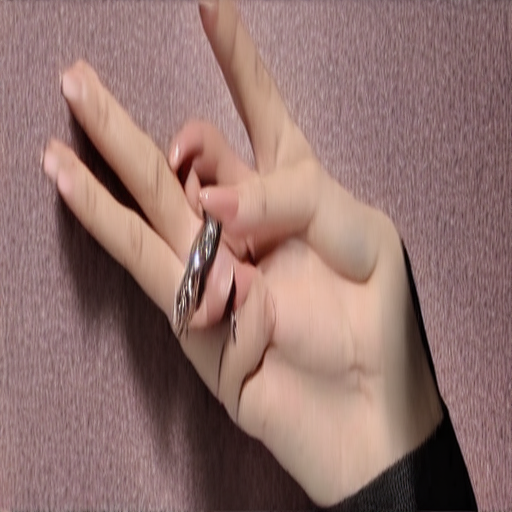} &
        \includegraphics[width=0.16\textwidth]{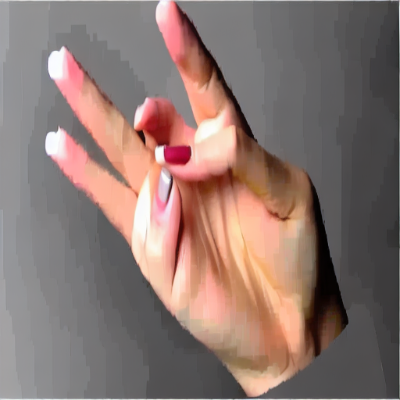}&
        \includegraphics[width=0.16\textwidth]{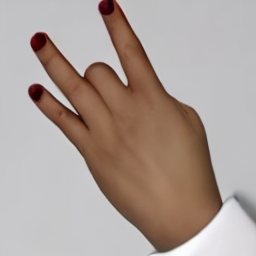} \\
        
        \includegraphics[width=0.16\textwidth]{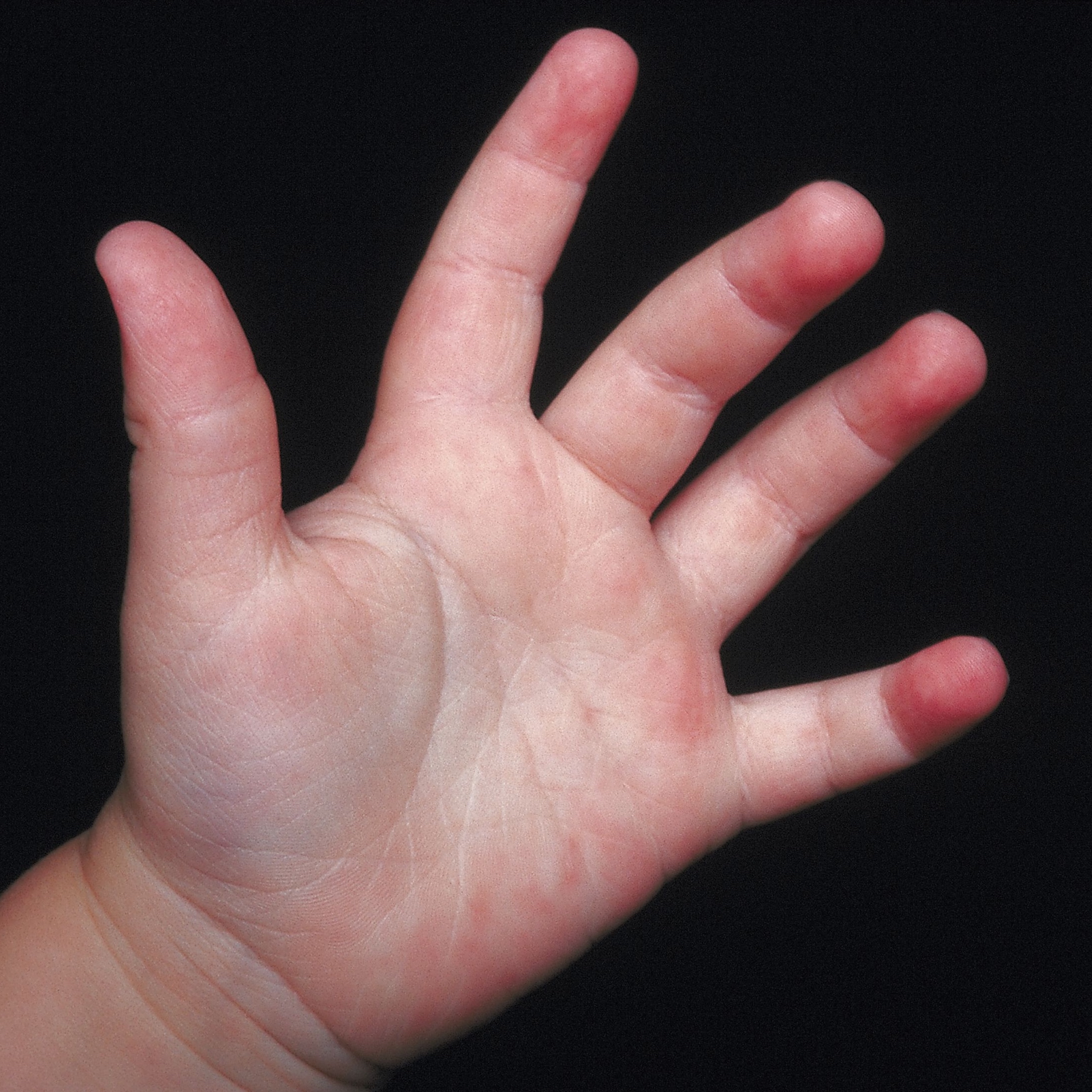}&
        \includegraphics[width=0.16\textwidth]{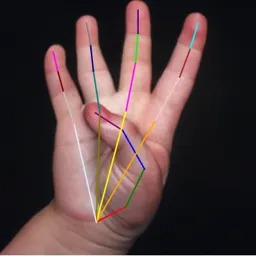}&
        \includegraphics[width=0.16\textwidth]{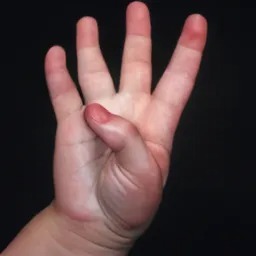}&
        \includegraphics[width=0.16\textwidth]{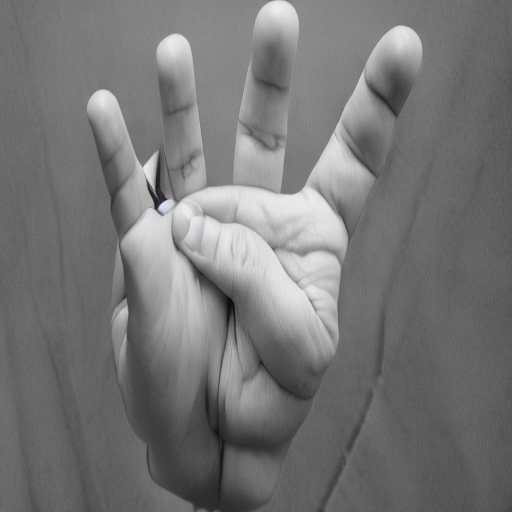} &
        \includegraphics[width=0.16\textwidth]{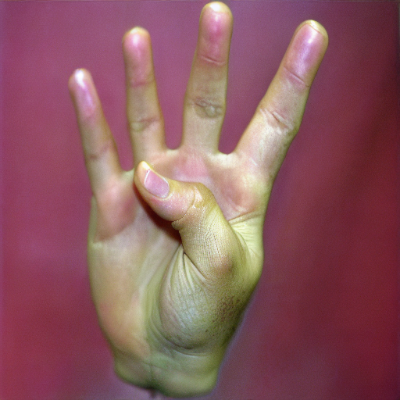}&
        \includegraphics[width=0.16\textwidth]{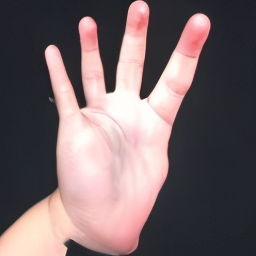} \\

        \includegraphics[width=0.16\textwidth]{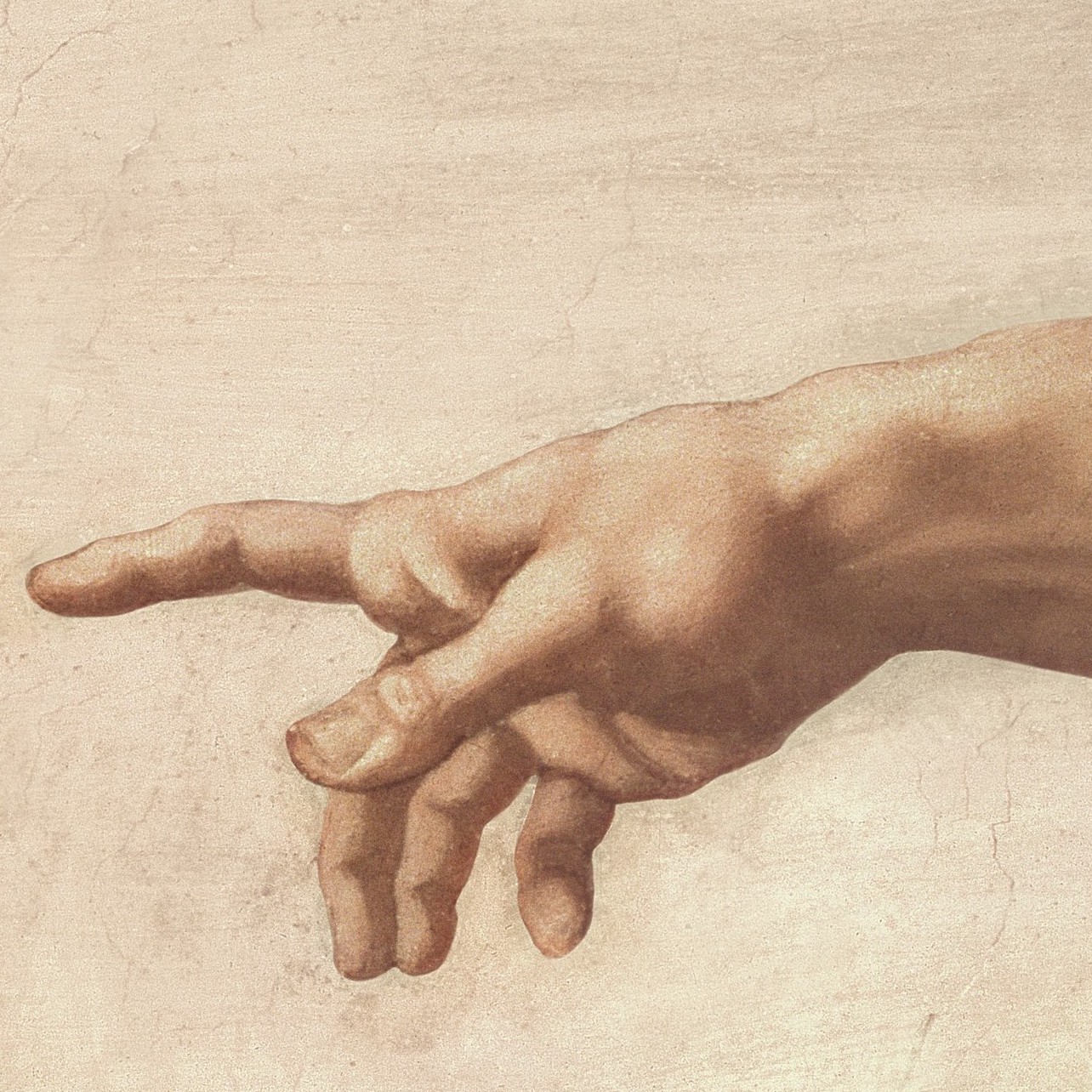}&
        \includegraphics[width=0.16\textwidth]{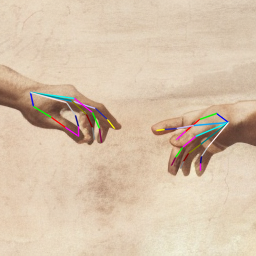}&
        \includegraphics[width=0.16\textwidth]{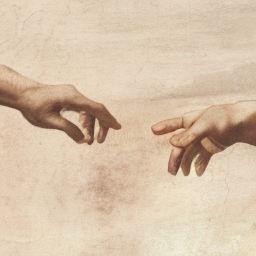}&
        \includegraphics[width=0.16\textwidth]{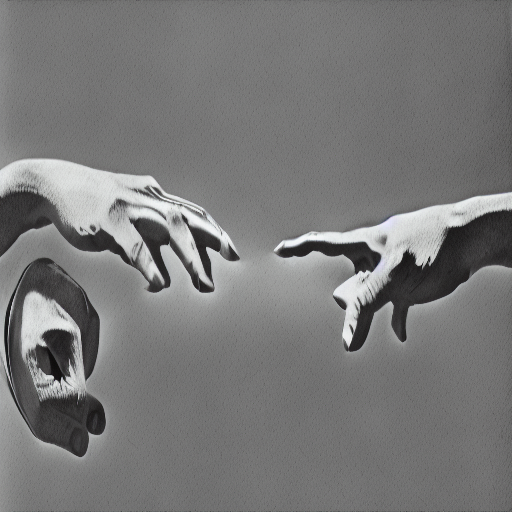} &
        \includegraphics[width=0.16\textwidth]{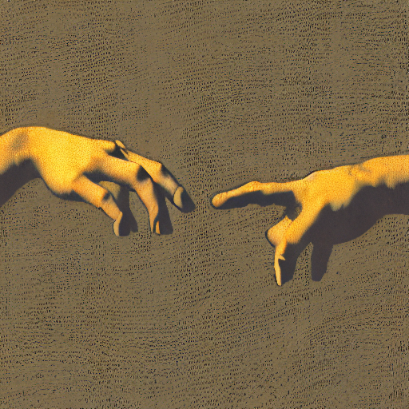}&
        \includegraphics[width=0.16\textwidth]{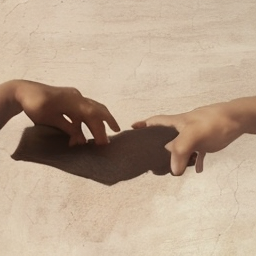} \\

        \includegraphics[width=0.16\textwidth]{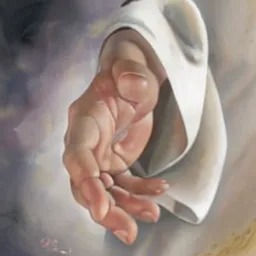}&
        \includegraphics[width=0.16\textwidth]{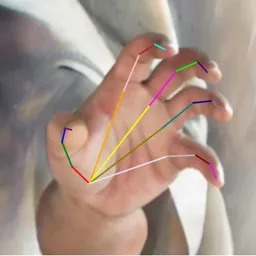}&
        \includegraphics[width=0.16\textwidth]{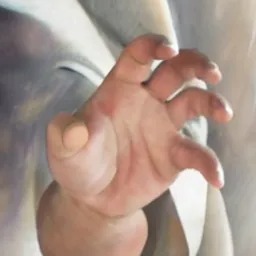}&
        \includegraphics[width=0.16\textwidth,height=0.16\textwidth]{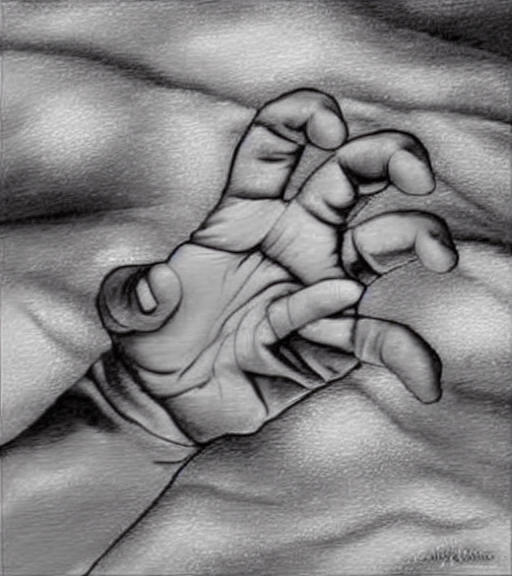} &
        \includegraphics[width=0.16\textwidth,height=0.16\textwidth]{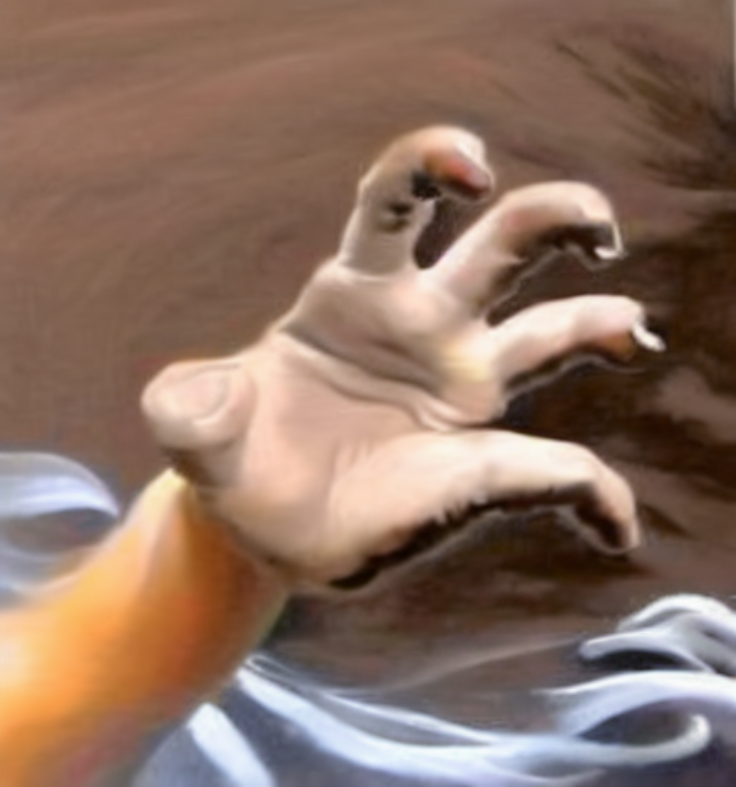} & 
        \includegraphics[width=0.16\textwidth]{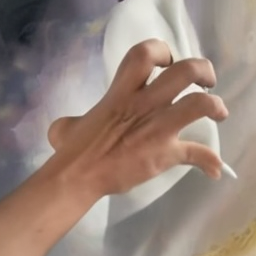}\\ [2.0pt]
        Reference & Ours w/ skeleton & Ours & Uni-ControlNet~\cite{qin2023unicontrol} & AnyControl~\cite{sun2024anycontrol} & CosHand~\cite{coshand}\\
    \end{tabular}
    \caption{Given reference images (top), our model transforms hands to target poses (visualized with ours at bottom) while faithfully preserving appearance details such as fingernails and textures. The model demonstrates generalization across diverse visual domains, from photorealistic images to artistic paintings, maintaining both anatomical plausibility and appearance fidelity.}
    \label{fig:gesture_transfer}
\end{figure*}

\section{Experiment}
\label{sec:experiment}
We test the generalization capability of our model across various 2D and 3D applications on in-the-wild web-sourced data and showcase that our versatile model outperforms state-of-the-art models.
We show more examples and videos in the supplement.

\paragraph{Gesture Transfer.}
We first evaluate our model's ability to transform hand poses while preserving image and hand appearance.
Given a reference hand image and a target pose (\eg,~extracted from another hand image), the goal is to generate a new image that maintains the reference appearance while adopting the target pose.
To demonstrate in-the-wild generalization, we curate a test set from web-sourced images with diverse styles.
In Fig.~\ref{fig:gesture_transfer} we compare \modelname with several baselines: GestureGAN~\cite{Tang2018GestureGANFH}, Uni-ControlNet~\cite{qin2023unicontrol}, AnyControl~\cite{sun2024anycontrol}, and CosHand~\cite{coshand}. Our model shows better fidelity than others. Moreover, it demonstrates faithful results in various configurations, including transformations between left and right hands, between single and dual hands, and artistic images.

For quantitative evaluation without ground truth targets, 
we measure reconstruction quality when the target pose matches the reference pose, where perfect preservation would yield identical images.
We assess these metrics using PSNR, SSIM, LPIPS, and FID scores between reference and generated images.

As seen in \Cref{tab:gesture_transfer}, \modelname significantly outperforms existing methods in simultaneously maintaining appearance and achieving accurate pose transfer.
While GestureGAN is specifically designed for gesture transfer, it exhibits significant overfitting to its training dataset.
Uni-ControlNet and AnyControl can take a reference image with depth maps and a skeleton image to specify the target hand pose, but still struggle to preserve reference appearance and follow challenging hand poses. 
CosHand generates hands and their physical interactions based on target hand masks, but it requires precise mask specification, which is impractical unless one knows the desired output in advance. 

\begin{table}[!tp]
    \centering \footnotesize
    \renewcommand{\tabcolsep}{7pt}
    \caption{Quantitative evaluation of \textbf{gesture transfer} using identity consistency metrics.
    Our model significantly outperforms both hand-specific methods and fine-tuned diffusion models with 3D depth conditioning, demonstrating superior ability in maintaining appearance while achieving accurate pose transfer.
    }    
    \begin{tabular}{ccccc}
        \hline
        & PSNR$\uparrow$ & SSIM$\uparrow$ & LPIPS$\downarrow$ & FID$\downarrow$\\
        \hline
        GestureGAN~\cite{Tang2018GestureGANFH} & 11.18 & 0.43 & 0.52 & 12.90\\
        Uni-ControlNet~\cite{qin2023unicontrol} & 9.41 & 0.32 & 0.48 & 11.01\\
        AnyControl~\cite{sun2024anycontrol} & 10.59 & 0.42 & 0.40 & 7.46\\
        CosHand~\cite{coshand} & 26.21 & 0.75 & 0.22 & 3.60\\
        \textbf{\modelname (Ours)} & \textbf{30.96} & \textbf{0.82} & \textbf{0.20} & \textbf{2.58}\\
        \hline
    \end{tabular}
    \label{tab:gesture_transfer}
    
\end{table}

\begin{figure}[!tp]
    \renewcommand{\tabcolsep}{0.0pt}
    \renewcommand{\arraystretch}{0.0}
    \centering \footnotesize
    \begin{tabular}{ccccc}
    \includegraphics[width=0.1\textwidth]{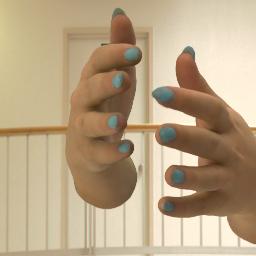} & 
    \includegraphics[width=0.1\textwidth]{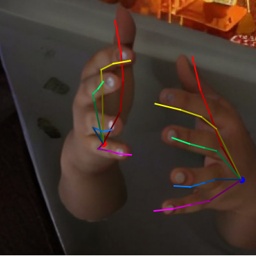} & 
    \includegraphics[width=0.1\textwidth]{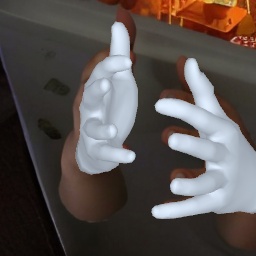} & 
    \includegraphics[width=0.1\textwidth]{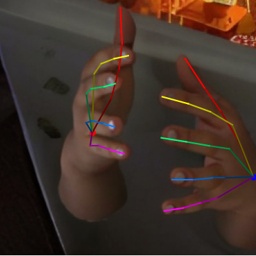} & 
    \includegraphics[width=0.1\textwidth]{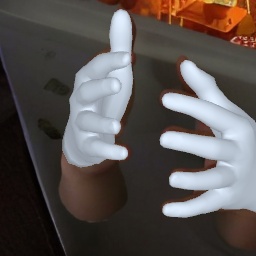} \\
    \includegraphics[width=0.1\textwidth]{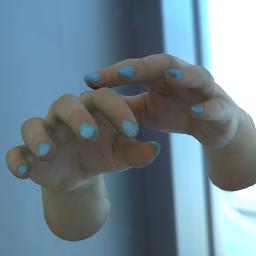} & 
    \includegraphics[width=0.1\textwidth]{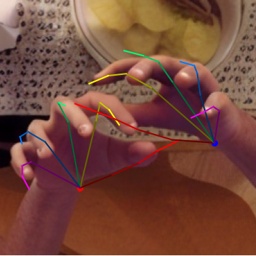} & 
    \includegraphics[width=0.1\textwidth]{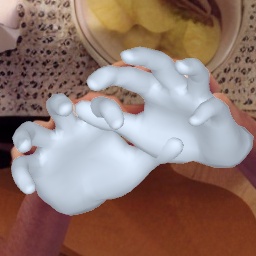} & 
    \includegraphics[width=0.1\textwidth]{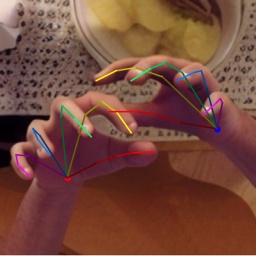} & 
    \includegraphics[width=0.1\textwidth]{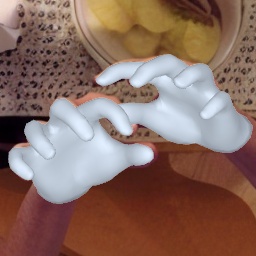} \\
    
    \includegraphics[width=0.1\textwidth]{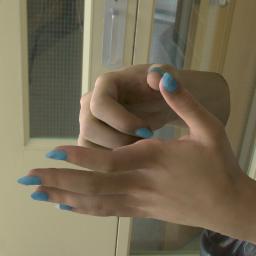} & 
    \includegraphics[width=0.1\textwidth]{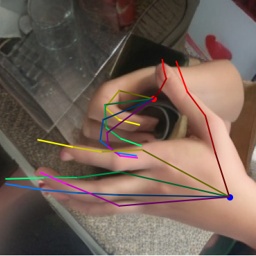} & 
    \includegraphics[width=0.1\textwidth]{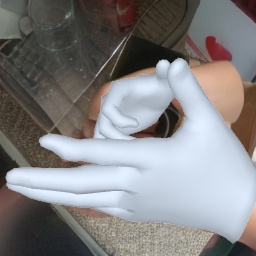} & 
    \includegraphics[width=0.1\textwidth]{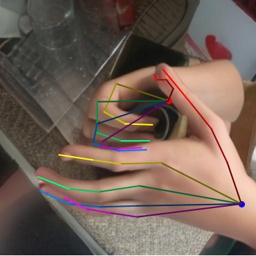} & 
    \includegraphics[width=0.1\textwidth]{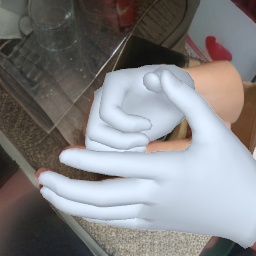} \\

    \includegraphics[width=0.1\textwidth]{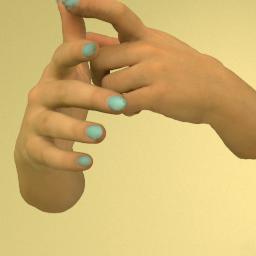} & 
    \includegraphics[width=0.1\textwidth]{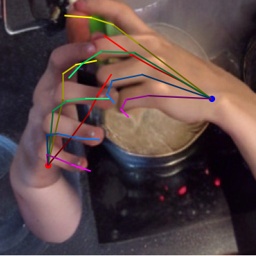} & 
    \includegraphics[width=0.1\textwidth]{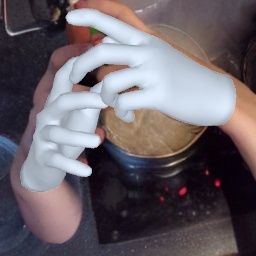} & 
    \includegraphics[width=0.1\textwidth]{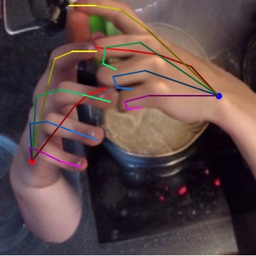} & 
    \includegraphics[width=0.1\textwidth]{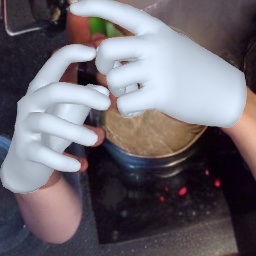} \\ 
    
    \vspace{0.3em}\\
    Synthetic data & \multicolumn{2}{c}{Before finetuning} & \multicolumn{2}{c}{After finetuning} \\
    \end{tabular}
    \caption{Given a synthetic hand dataset, \modelname can transform it to the in-the-wild domain with realistic appearance and background, improving existing 3D hand estimation after finetuning on our generated data.}
    \label{fig:domain_transfer}
\end{figure}

\paragraph{Domain Transfer.}
Well-annotated hand datasets often come from controlled environments (laboratories, studios) or synthetic data, with distributions different from real-world data.
\modelname can address this limitation by transforming hand images from a source domain to a target in-the-wild domain using only reference images.
To demonstrate this, we transfer 10K hand images from ReInterHand~\cite{reinterhand} to the EpicKitchen~\cite{epickitchen} domain.
The evaluation (\Cref{tab:domain_transfer} and \Cref{fig:domain_transfer}) shows that fine-tuning a state-of-the-art 3D hand estimator~\cite{hamer} on small set of transformed data further improves performance.
These results (\Cref{fig:domain_transfer}), establish \modelname as an effective data augmentation tool for enhancing existing hand understanding models.

\begin{figure}[!tp]
    \renewcommand{\tabcolsep}{0.0pt}
    \renewcommand{\arraystretch}{0.}
    \centering \footnotesize
    \begin{tabular}{ccccccc}
    \includegraphics[width=0.095\textwidth]{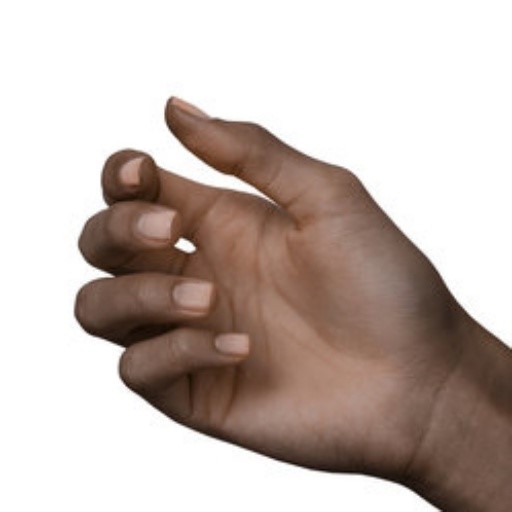} &
    $\;$&
    \includegraphics[width=0.095\textwidth]{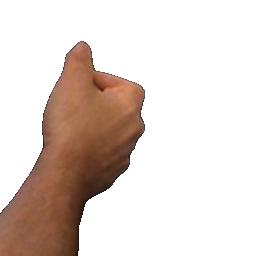} &
    \includegraphics[width=0.095\textwidth]{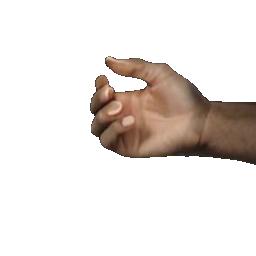} &
    \includegraphics[width=0.095\textwidth]{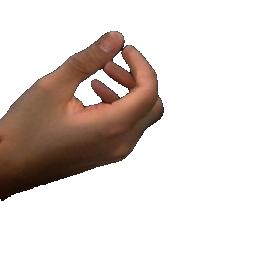} &
    \includegraphics[width=0.095\textwidth]{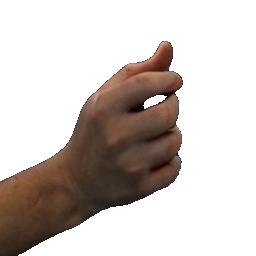} \\
    
    \includegraphics[width=0.095\textwidth]{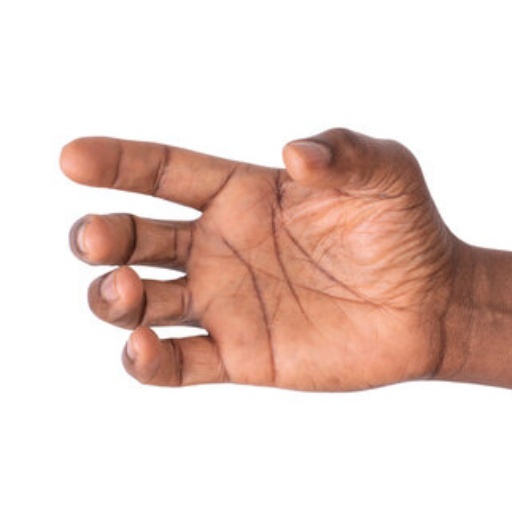} & 
    &
    \includegraphics[width=0.095\textwidth]{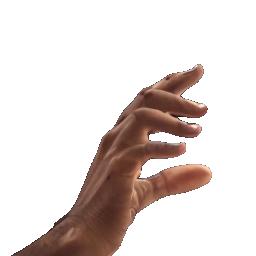} & 
    \includegraphics[width=0.095\textwidth]{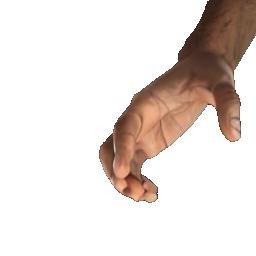} & 
    \includegraphics[width=0.095\textwidth]{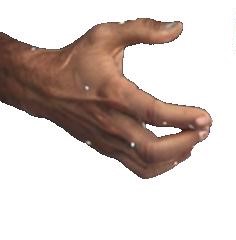} & 
    \includegraphics[width=0.09\textwidth]{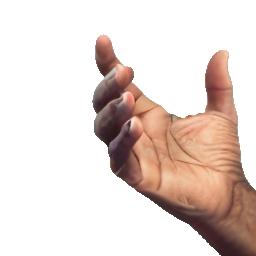} \\
    
    \includegraphics[width=0.095\textwidth]{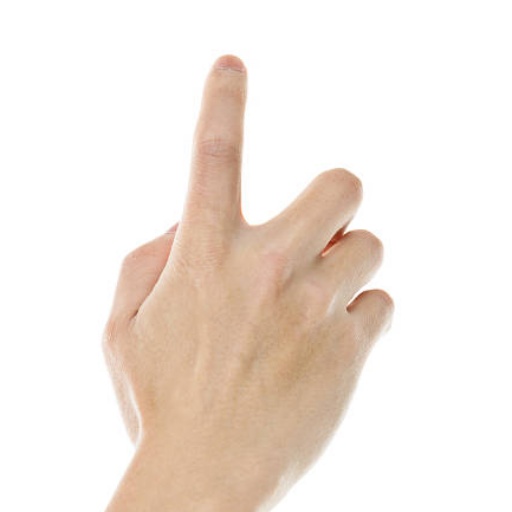} & 
    &
    \includegraphics[width=0.095\textwidth]{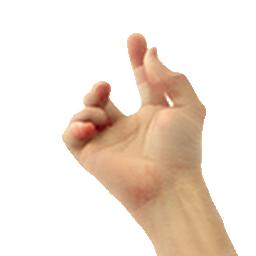} &
    \includegraphics[width=0.095\textwidth]{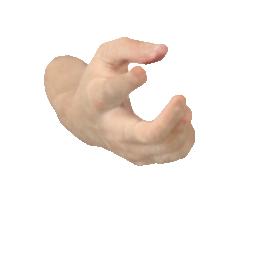} & 
    \includegraphics[width=0.095\textwidth]{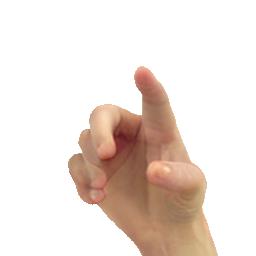} & 
    \includegraphics[width=0.095\textwidth]{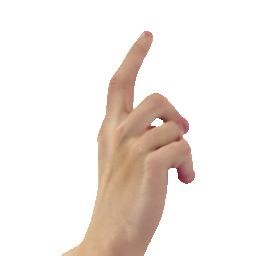} \\[3.0pt]
    Reference & & View \#1 & View \#2 & View \#3 & View \#4 \\
    \end{tabular}
    \caption{From a single input image (1st column), \modelname generates diverse viewpoints, demonstrating robust generalization to unseen hands and camera poses.}
    \vspace{0.2in}
    \label{fig:nvs_itw}
    \renewcommand{\tabcolsep}{0pt}
    \renewcommand{\arraystretch}{0}
    \centering \footnotesize
    \begin{tabular}{cccc}
        \includegraphics[width=0.12\textwidth]{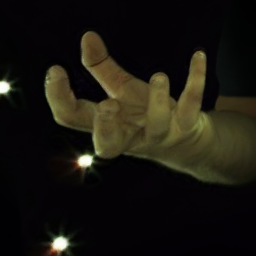} &
        \includegraphics[width=0.12\textwidth]{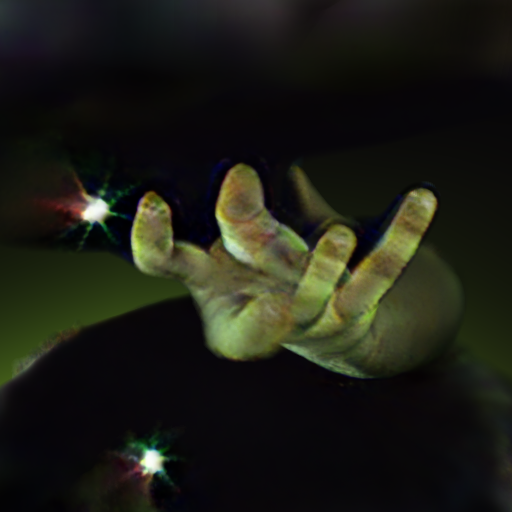} &
        \includegraphics[width=0.12\textwidth]{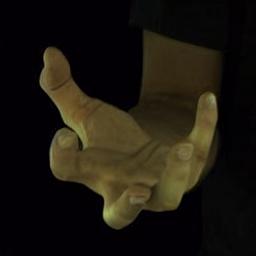} &
        \includegraphics[width=0.12\textwidth]{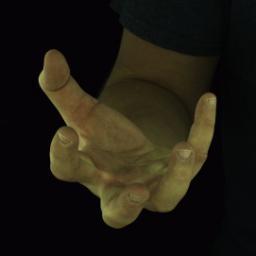} \\
        \includegraphics[width=0.12\textwidth]{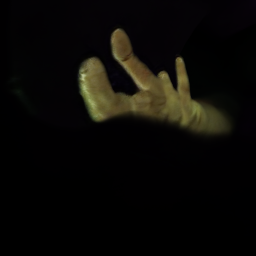} &
        \includegraphics[width=0.12\textwidth]{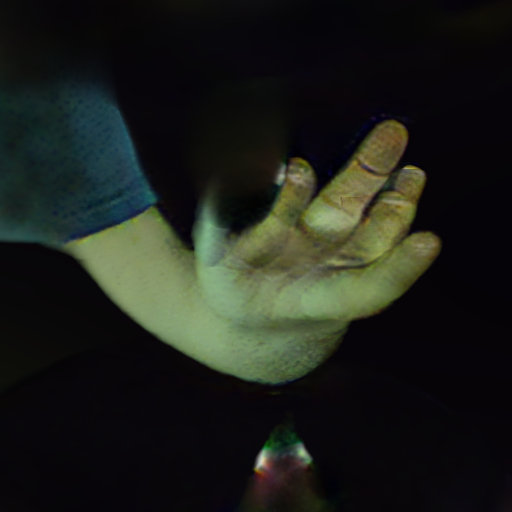} &
        \includegraphics[width=0.12\textwidth]{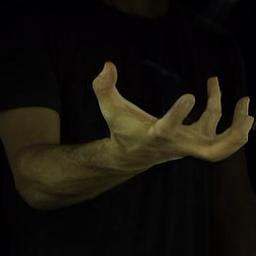} &
        \includegraphics[width=0.12\textwidth]{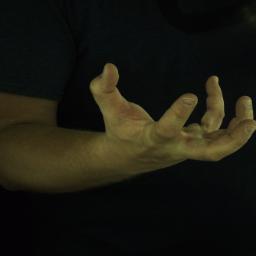} \\
        \includegraphics[width=0.12\textwidth]{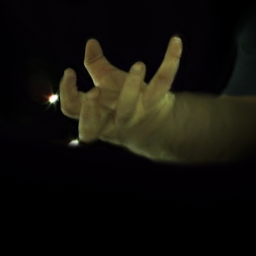} &
        \includegraphics[width=0.12\textwidth]{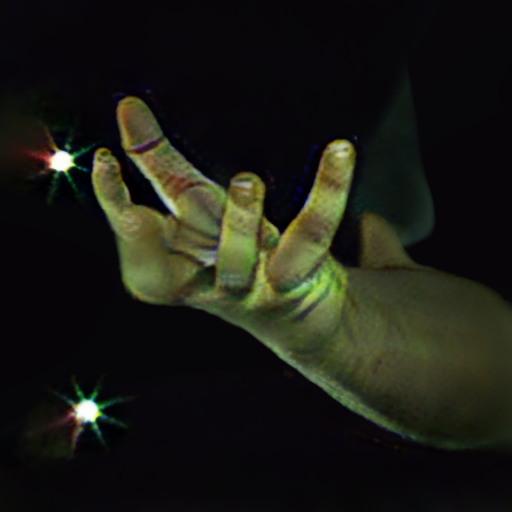} &
        \includegraphics[width=0.12\textwidth]{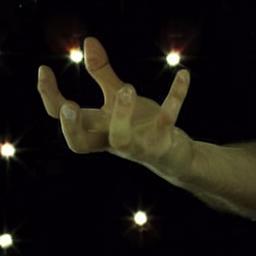} &
        \includegraphics[width=0.12\textwidth]{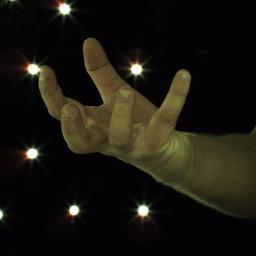} \\[3.0pt]
        ZeroNVS\cite{zeronvs} & ImageDream\cite{imagedream} & Ours & GT (unseen) \\
    \end{tabular}
    \caption{Novel view synthesis on InterHand2.6M~\cite{interhand}.}
    \label{fig:nvs}
\end{figure}

\begin{table}[t]
    \centering \footnotesize
    \caption{After fine-tuning HaMeR~\cite{hamer} on our domain transferred data synthesized by \modelname, we can further improve performance of the 3D mesh recovery on the target domain.}
    \begin{tabular}{ccccc}
        \toprule
        & {\footnotesize PA-MPJPE$\downarrow$} & {\footnotesize PA-MPVPE$\downarrow$} & {\footnotesize F@5$\uparrow$} & {\footnotesize F@15$\uparrow$}\\
        \midrule
        Before finetune & 14.41 & 7.52 & 0.30 & 0.93\\
        After finetune& \textbf{13.68} & \textbf{6.29} & \textbf{0.46} & \textbf{0.96} \\
        \bottomrule
    \end{tabular}
    \label{tab:domain_transfer}
\vspace{0.5em}
    \centering \footnotesize
    \caption{(a) Novel view synthesis evaluation on InterHand2.6M~\cite{interhand}. \modelname outperforms methods leveraging explicit 3D representation despite using only learned 2D priors. (b) \modelname achieves competitive performance compared to pose-conditioned video diffusion models, despite being trained only on image pairs rather than video sequences.
    } 
    \begin{tabular}{ccccc}
        \toprule
        && PSNR$\uparrow$ & SSIM$\uparrow$ & LPIPS$\downarrow$\\
        \midrule
        &Zero123~\cite{liu2023zero1to3} & 11.37 & 0.79 & 0.17\\
        (a) NVS&ZeroNVS~\cite{zeronvs} & 19.21 &	0.74 & 0.24 \\
        &ImageDream~\cite{imagedream} & 19.97 & 0.80 & 0.17\\
        &\textbf{\modelname (Ours)} & \textbf{27.72} & \textbf{0.88}	& \textbf{0.10}\\
        \midrule
        &ControlNeXt~\cite{controlnext}& 17.64	& 0.73 &	0.29\\
        (b) Video&AnimateAnyone~\cite{animate_anyone} & 15.76	&0.74	&0.35\\
        &\textbf{\modelname (Ours)} & \textbf{24.08} & \textbf{0.83} & \textbf{0.17}\\
        \bottomrule
    \end{tabular}
    \label{tab:nvs_video}
\end{table}

\vspace{-0.15in}
\paragraph{Novel View Synthesis (NVS).}
Novel view synthesis for hands presents unique challenges due to their complex articulation and significant appearance variations across viewpoints.
We evaluate our approach against state-of-the-art view synthesis methods Zero123~\cite{zero123++}, ZeroNVS~\cite{zeronvs}, and ImageDream~\cite{imagedream} on the InterHand2.6M dataset.
While ZeroNVS and ImageDream ensure 3D consistency through explicit NeRF~\cite{nerf} representations and with costly test-time SDS~\cite{dreamfusion} loss (\cite{zeronvs,imagedream} takes 5 hours and 80GB VRAM for each instance), \modelname achieves superior performance through learned hand-specific priors without explicit 3D constraints.
Results in InterHand2.6M (\Cref{tab:nvs_video} and \Cref{fig:nvs}) demonstrate that our model outperforms the baselines by 39\% higher PSNR, and \Cref{fig:nvs_itw} shows our model's impressive generalization to in-the-wild web-sourced images.

%% file: sec/5_applications.tex
\begin{figure*}[!tp]
    \renewcommand{\tabcolsep}{0pt}
    \renewcommand{\arraystretch}{0.}
    \centering \footnotesize
    \begin{tabular}{ccccccccc}
        \rotatebox{90}{\hspace{0.5cm} Malformed}&
        \; &
        \includegraphics[width=0.14\textwidth]{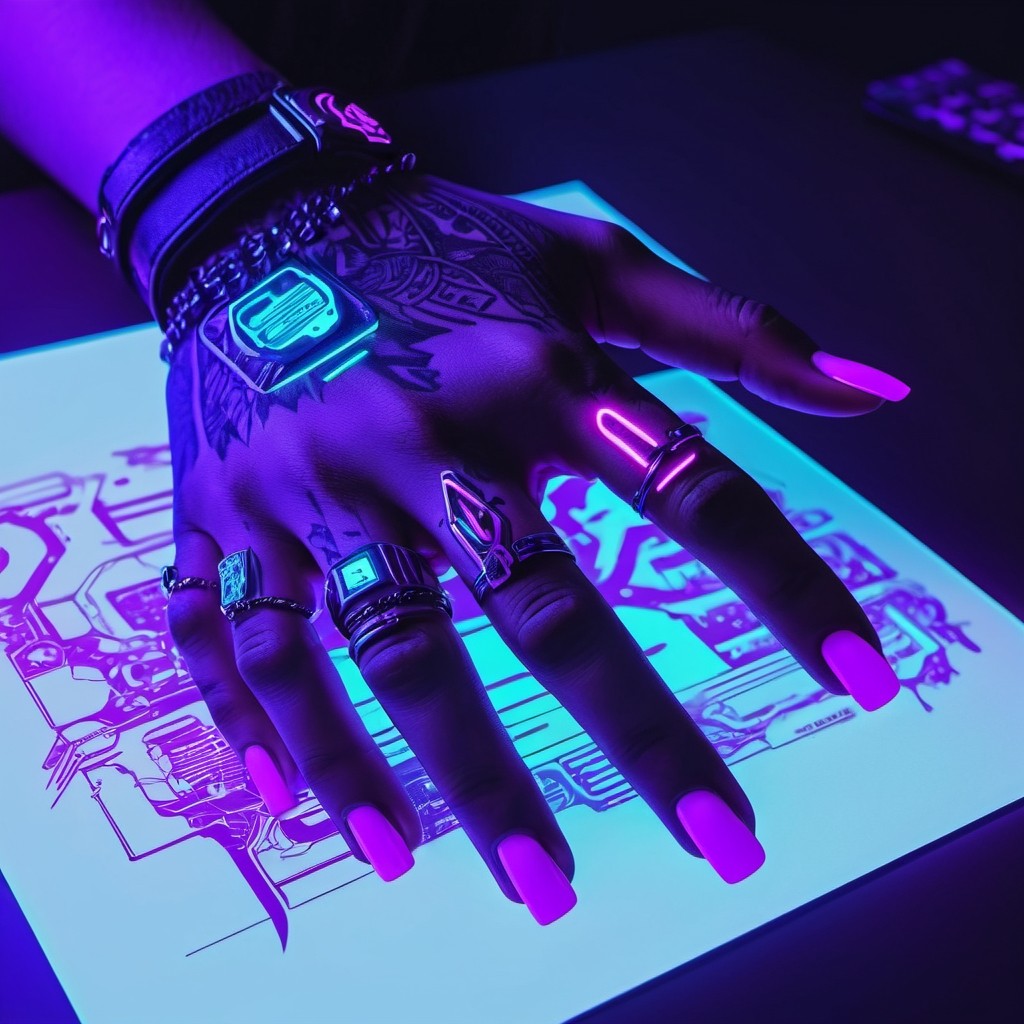} & 
        \includegraphics[width=0.14\textwidth]{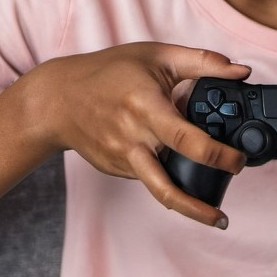} & 
        \includegraphics[width=0.14\textwidth]{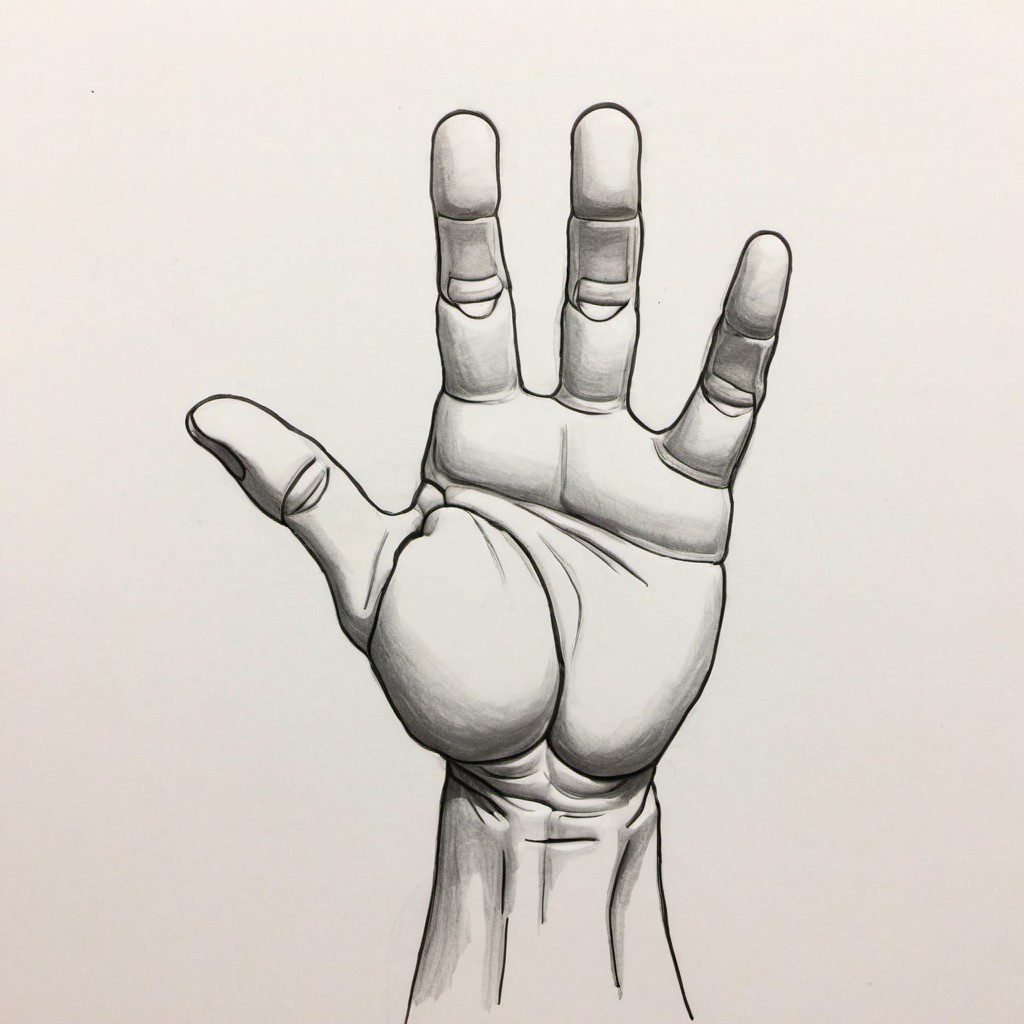} & 
        \includegraphics[width=0.14\textwidth]{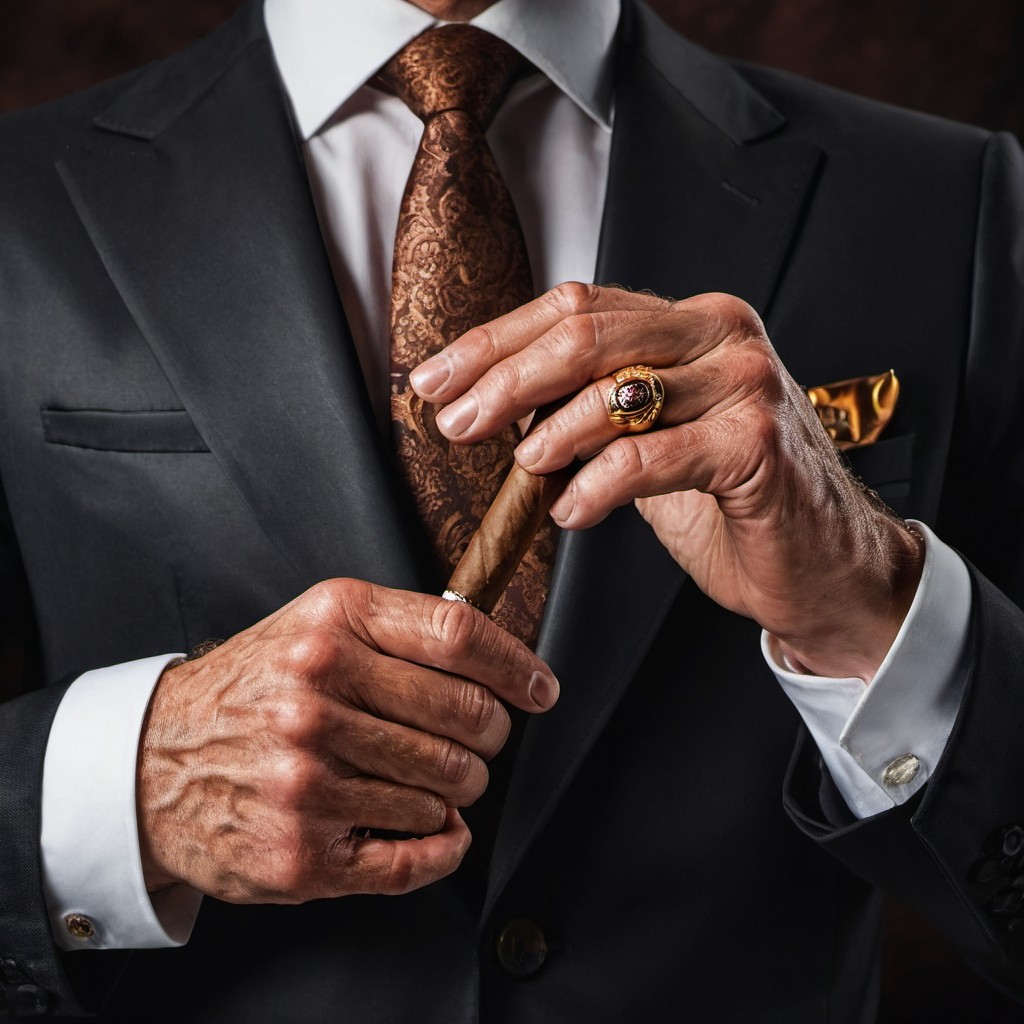} & 
        \includegraphics[width=0.14\textwidth]{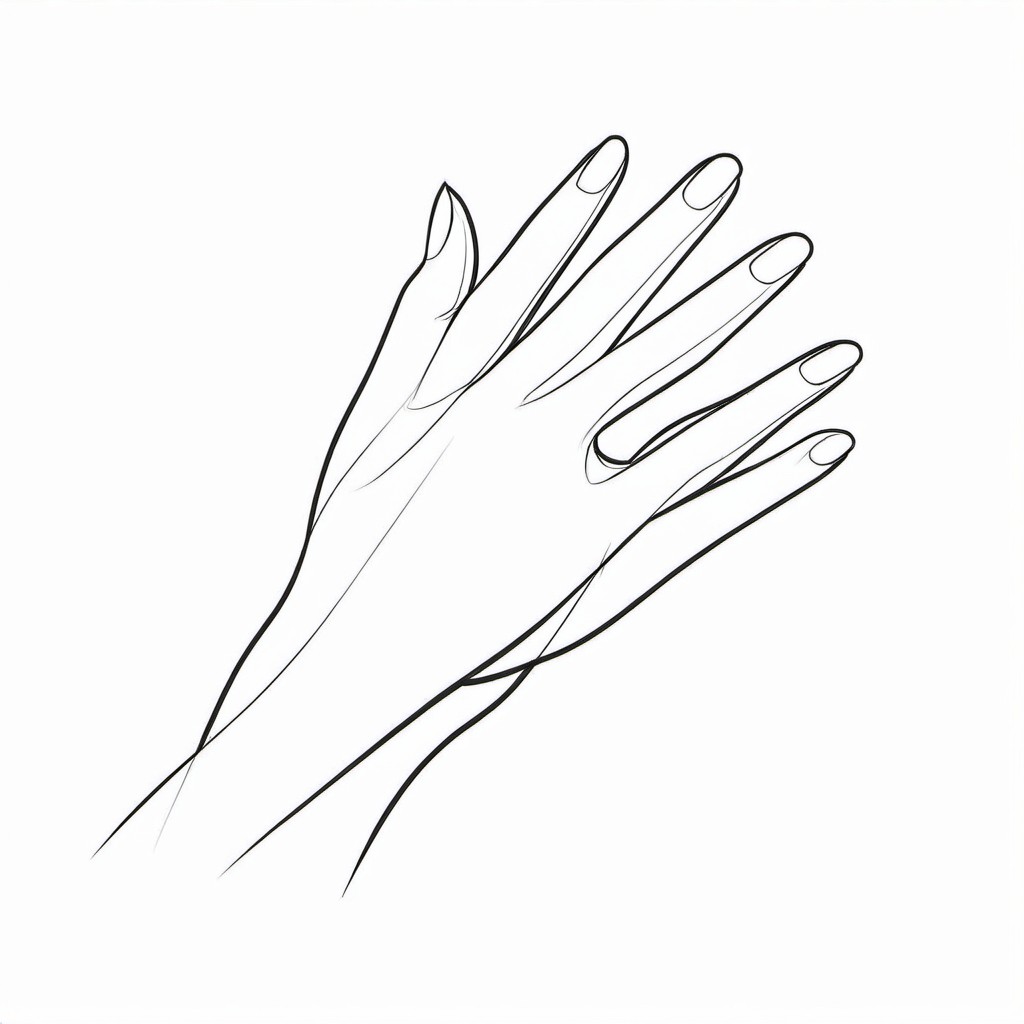} & 
        \includegraphics[width=0.14\textwidth]{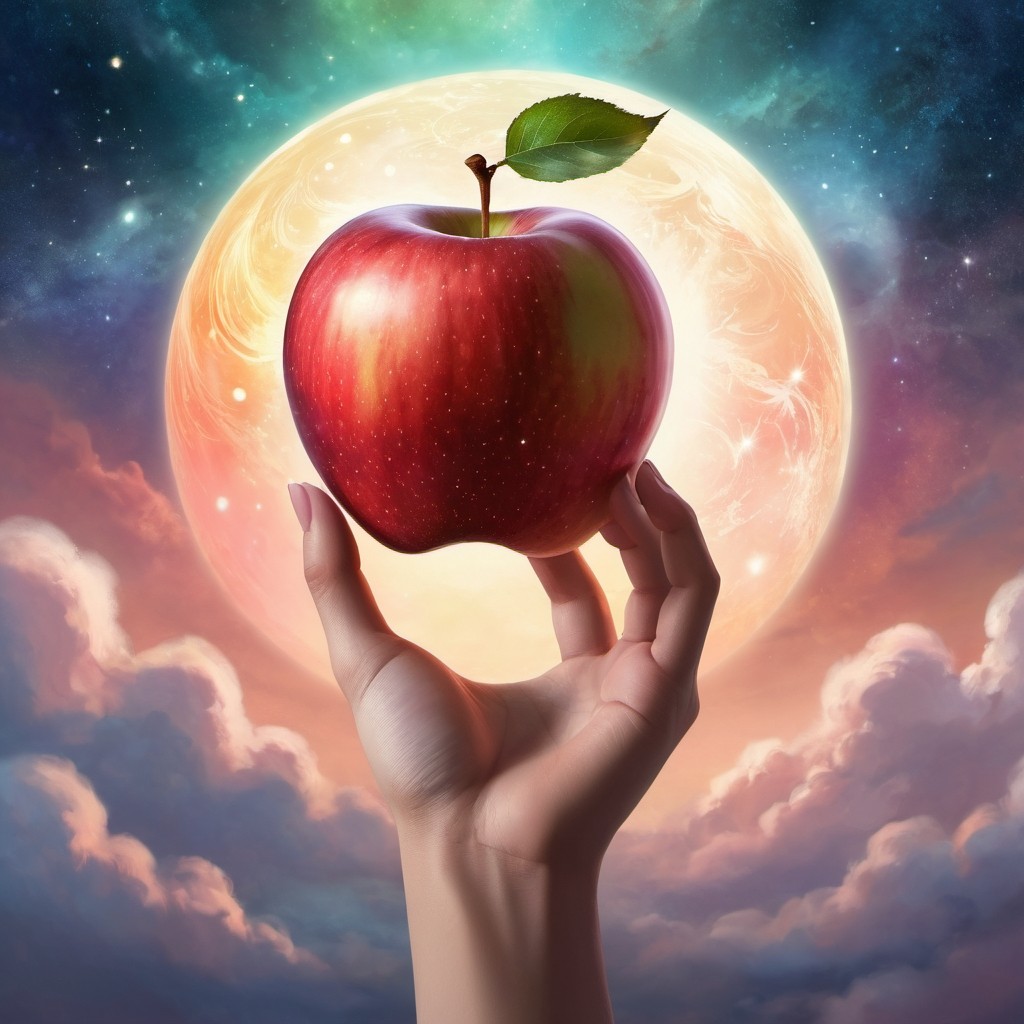} & 
        \includegraphics[width=0.14\textwidth]{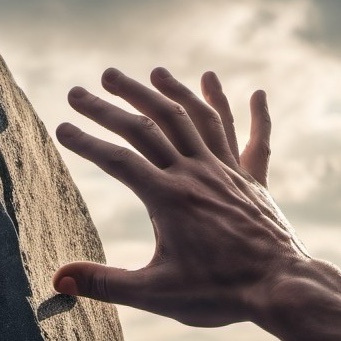} \\ 
         \rotatebox{90}{\hspace{1cm} \cite{lu2023handrefiner}}&
         &
        \includegraphics[width=0.14\textwidth]{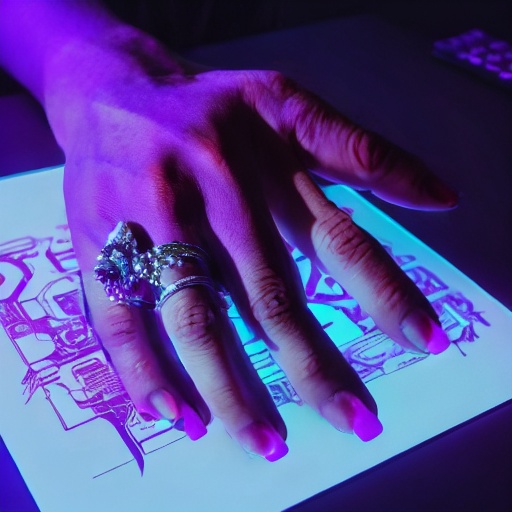} & 
        \includegraphics[width=0.14\textwidth]{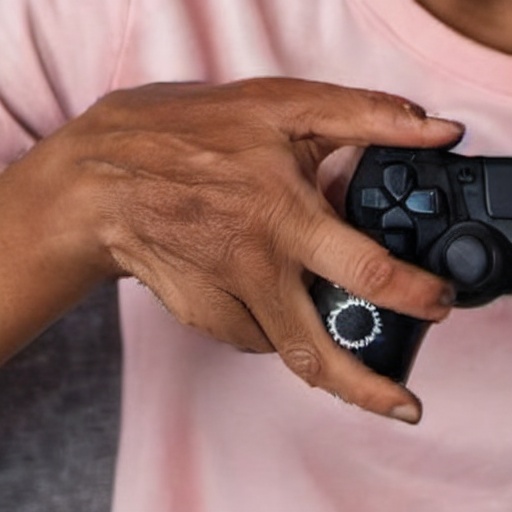} & 
        \includegraphics[width=0.14\textwidth]{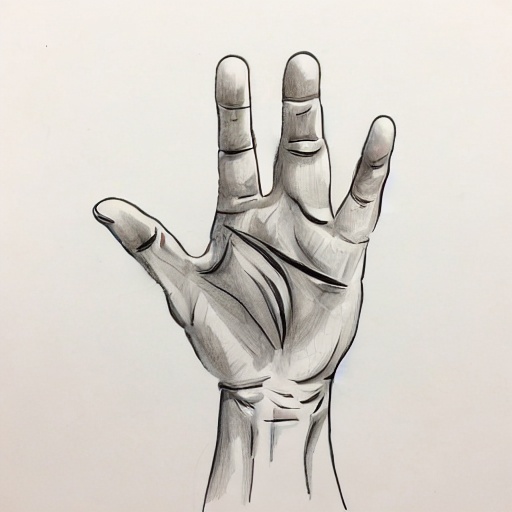} & 
        \includegraphics[width=0.14\textwidth]{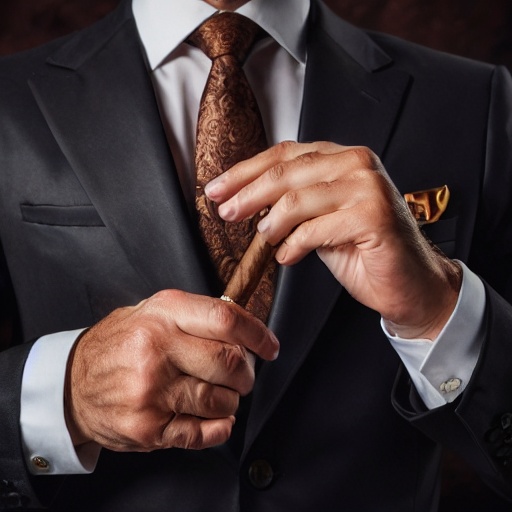} & 
        \includegraphics[width=0.14\textwidth]{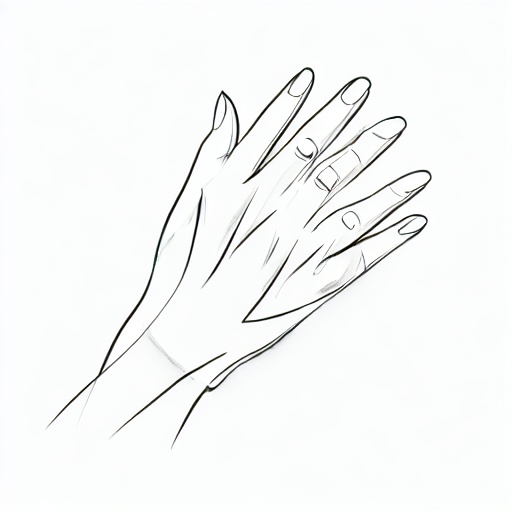} & 
        \includegraphics[width=0.14\textwidth]{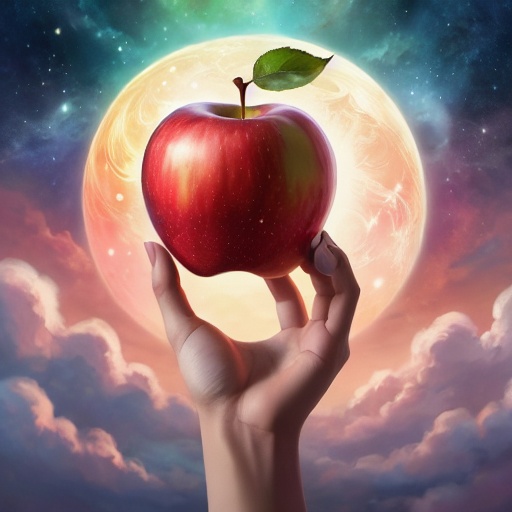} & 
        \includegraphics[width=0.14\textwidth]{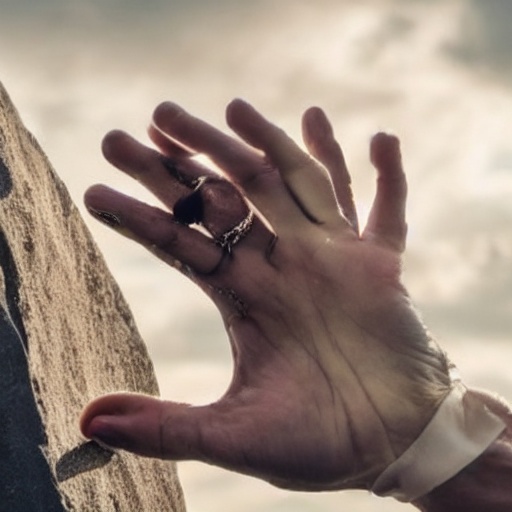} \\ 
        \rotatebox{90}{\hspace{1cm} \cite{realishuman}}&
        &
        \includegraphics[width=0.14\textwidth]{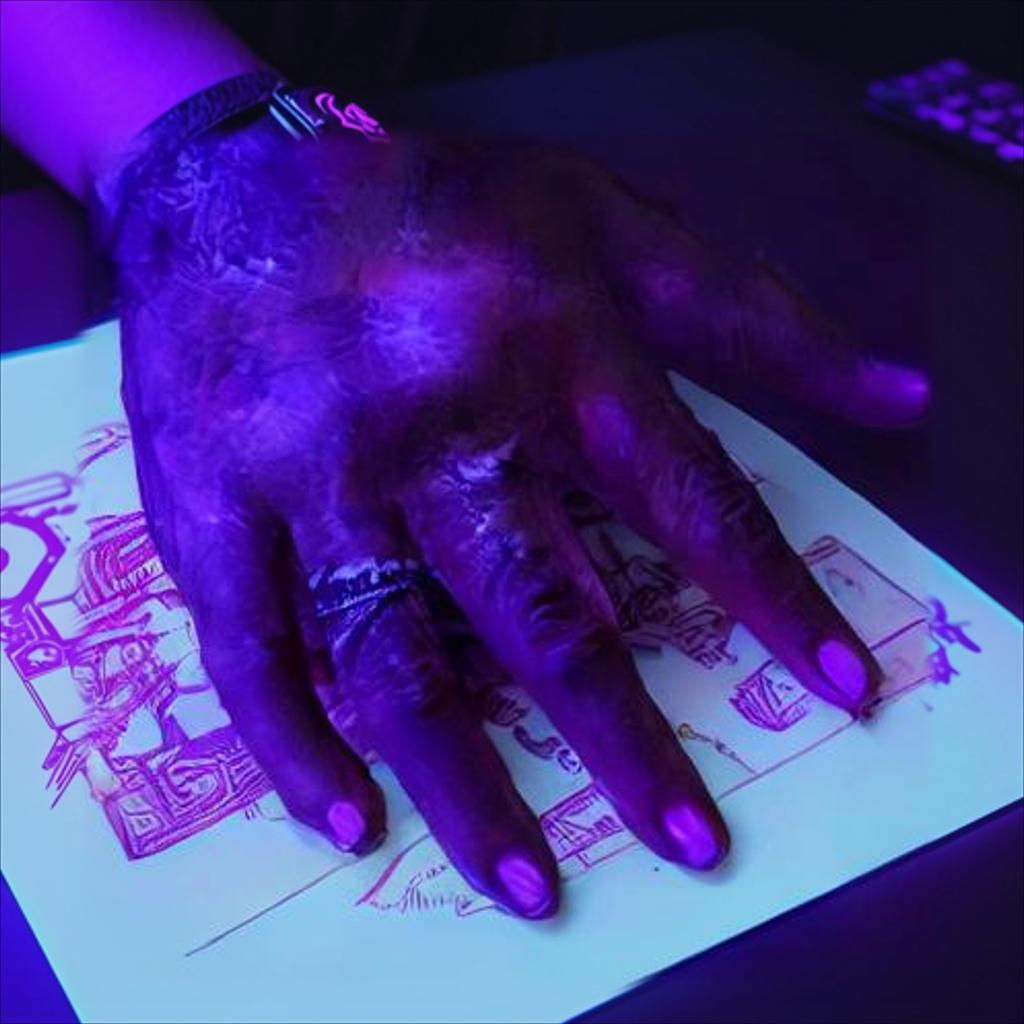} & 
        \includegraphics[width=0.14\textwidth]{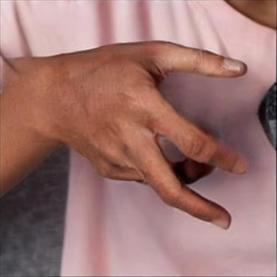} & 
        \includegraphics[width=0.14\textwidth]{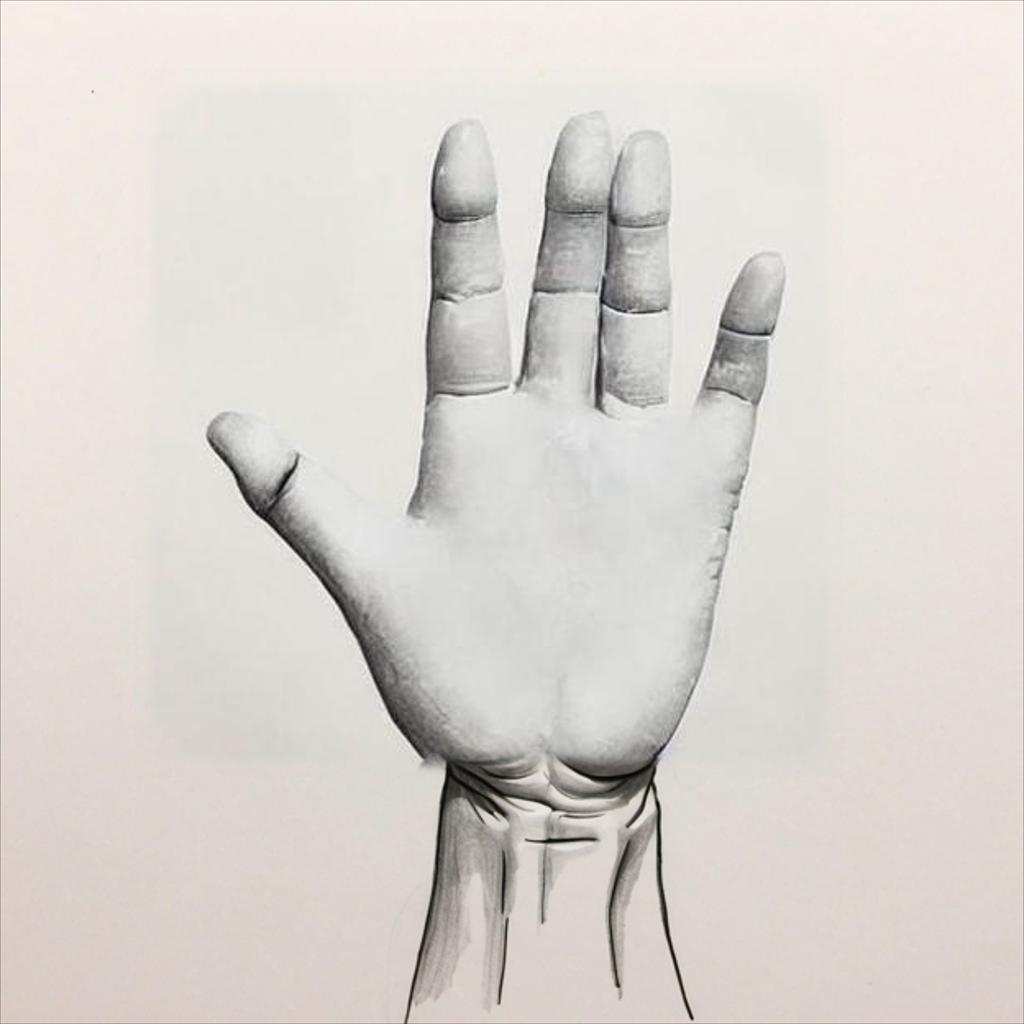} & 
        \includegraphics[width=0.14\textwidth]{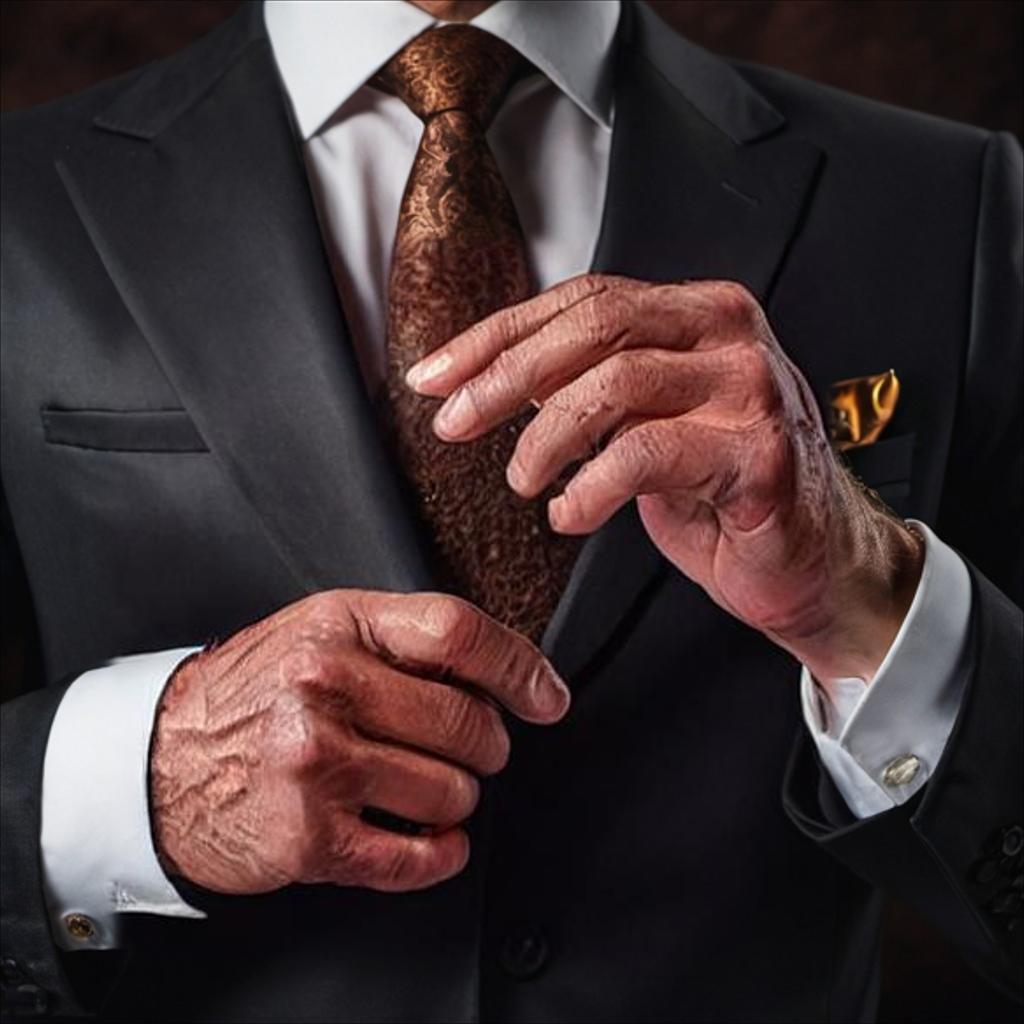} & 
        \includegraphics[width=0.14\textwidth]{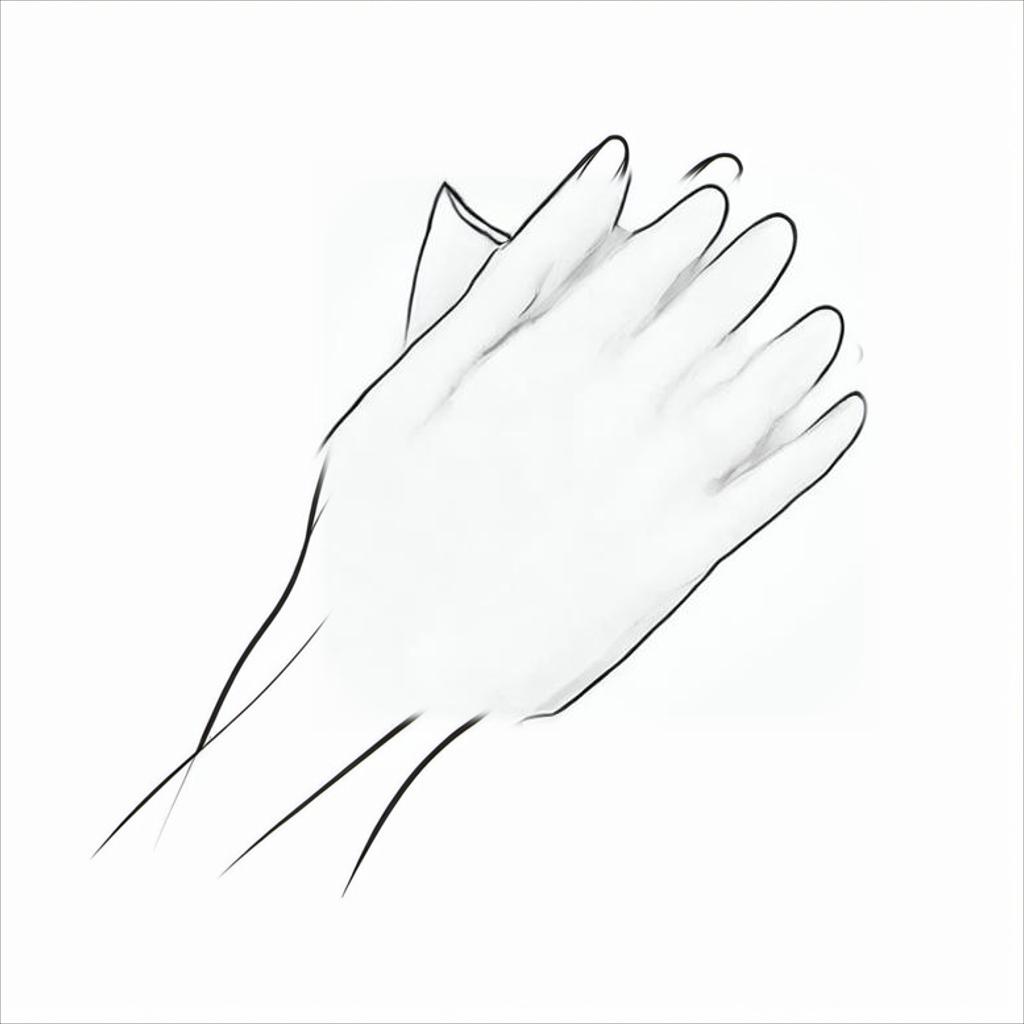} & 
        \includegraphics[width=0.14\textwidth]{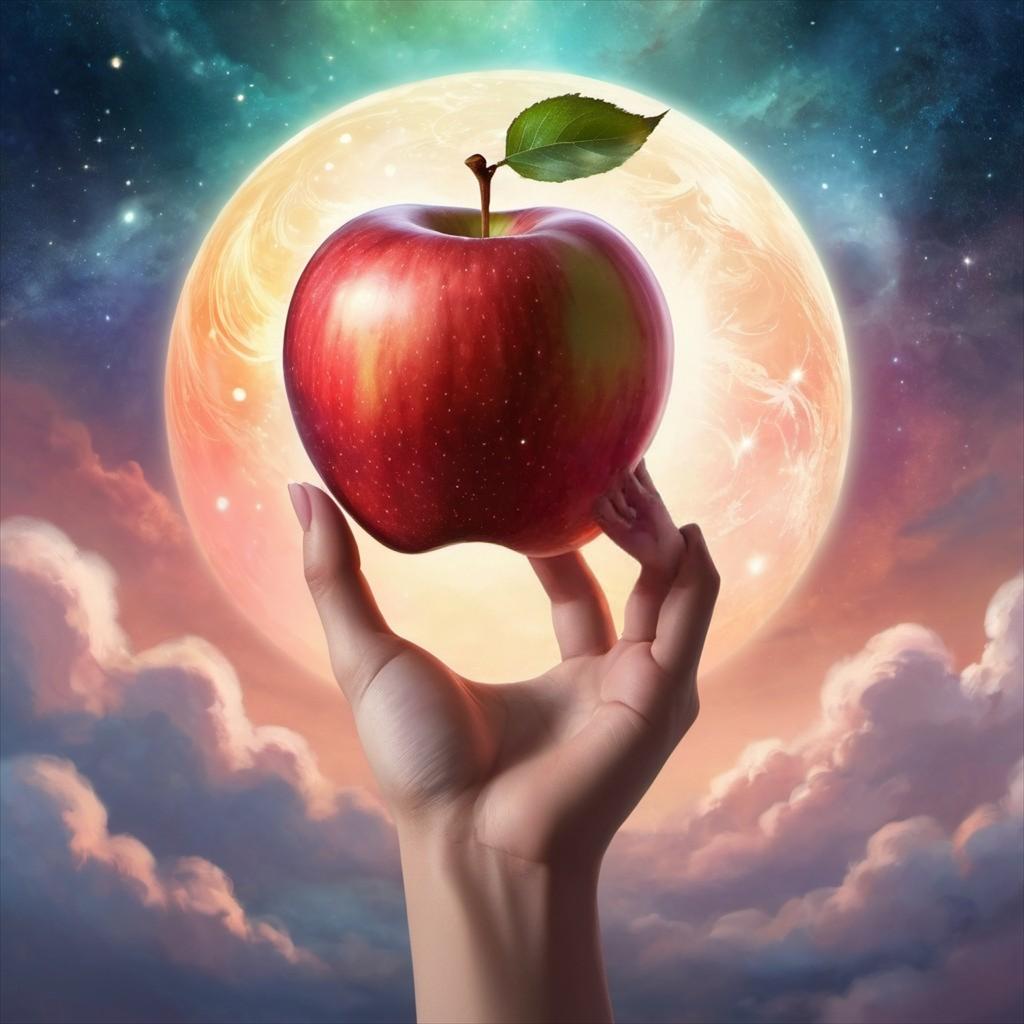} & 
        \includegraphics[width=0.14\textwidth]{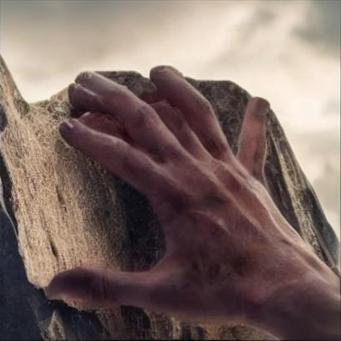} \\ 
        \rotatebox{90}{\hspace{1cm} Ours}&
        &
        \includegraphics[width=0.14\textwidth]{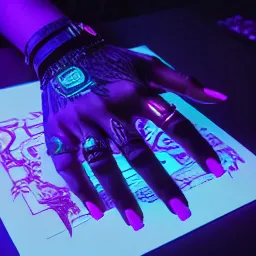}&
        \includegraphics[width=0.14\textwidth]{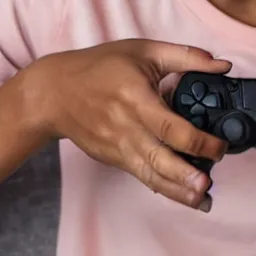}&
        \includegraphics[width=0.14\textwidth]{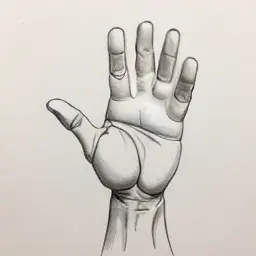}&
        \includegraphics[width=0.14\textwidth]{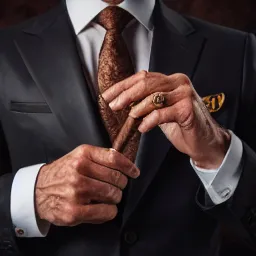}&
        \includegraphics[width=0.14\textwidth]{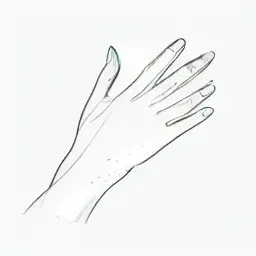}&
        \includegraphics[width=0.14\textwidth]{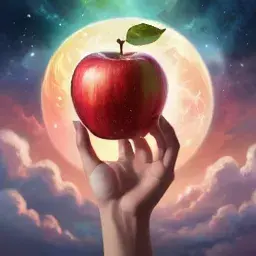}&
        \includegraphics[width=0.14\textwidth]{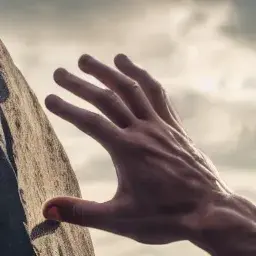}\\
    \end{tabular}
    \caption{Our \modelname develops zero-shot hand-fixing ability with exceptional preservation of the appearance of the hand as well as the context of hand-object interaction, compared with specialized hand fixer methods.}
    \label{fig:handfixer}
\end{figure*}

\begin{figure}[!tp]
    \renewcommand{\tabcolsep}{0pt}
    \renewcommand{\arraystretch}{0}
    \centering \footnotesize
    \begin{tabular}{cccccccc}
    \rotatebox{90}{\hspace{0.5cm} \footnotesize\cite{coshand}} & 
    \; &
    \includegraphics[width=0.09\textwidth]{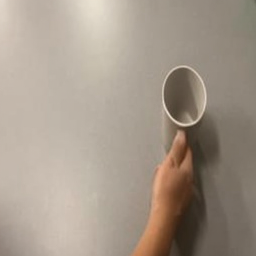} & 
    \includegraphics[width=0.09\textwidth]{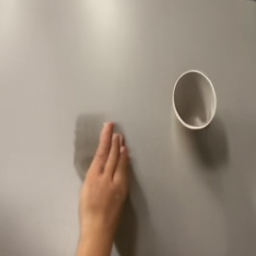} & 
    \includegraphics[width=0.09\textwidth]{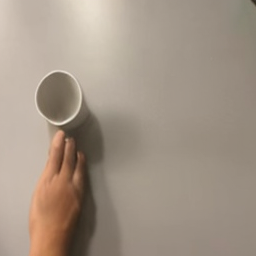} & 
    \includegraphics[width=0.09\textwidth]{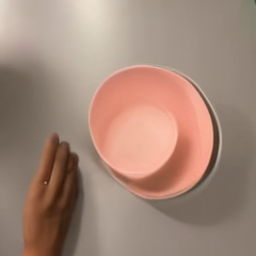} & 
    \includegraphics[width=0.09\textwidth]{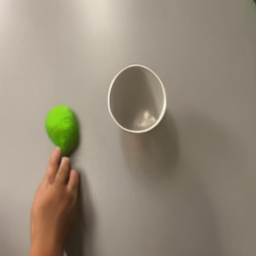} \\
    \rotatebox{90}{\hspace{0.5cm}\footnotesize Ours}&
    &
    \includegraphics[width=0.09\textwidth]{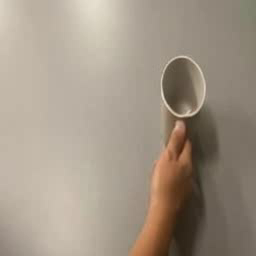} & 
    \includegraphics[width=0.09\textwidth]{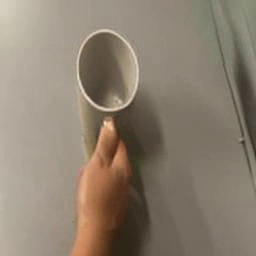} & 
    \includegraphics[width=0.09\textwidth]{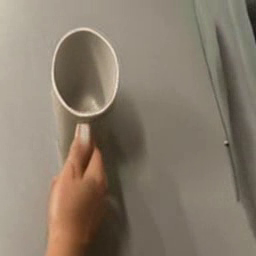} & 
    \includegraphics[width=0.09\textwidth]{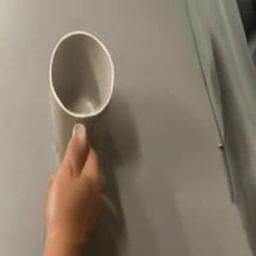} & 
    \includegraphics[width=0.09\textwidth]{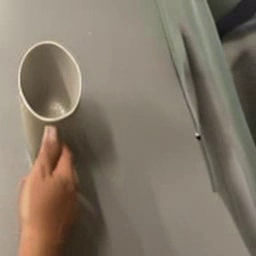} \\
    \rotatebox{90}{\hspace{0.5cm}\footnotesize\cite{coshand}}&
    &
    \includegraphics[width=0.09\textwidth]{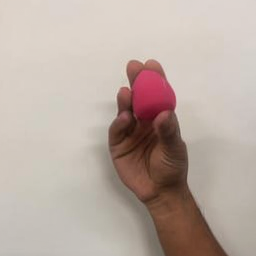} & 
    \includegraphics[width=0.09\textwidth]{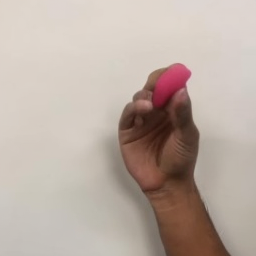} & 
    \includegraphics[width=0.09\textwidth]{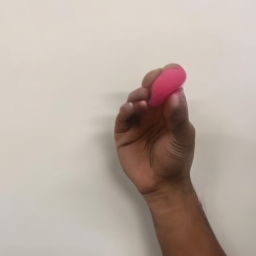} & 
    \includegraphics[width=0.09\textwidth]{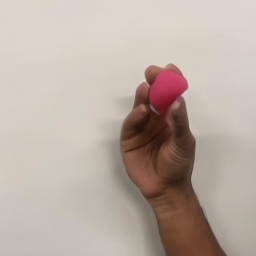} & 
    \includegraphics[width=0.09\textwidth]{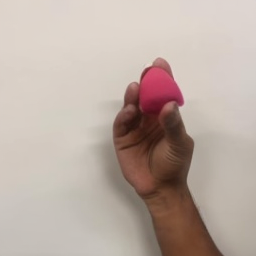} \\
    \rotatebox{90}{\hspace{0.5cm}\footnotesize Ours}&
    &
    \includegraphics[width=0.09\textwidth]{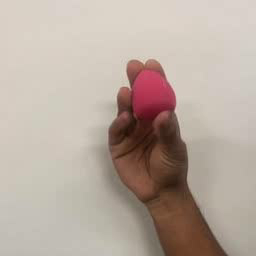} & 
    \includegraphics[width=0.09\textwidth]{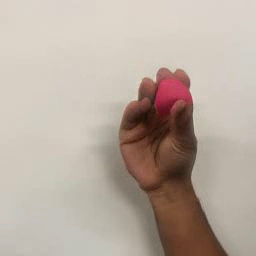} & 
    \includegraphics[width=0.09\textwidth]{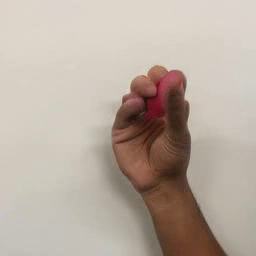} & 
    \includegraphics[width=0.09\textwidth]{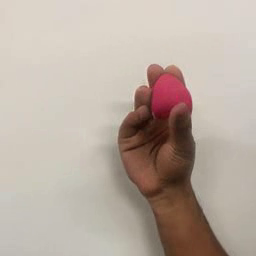} & 
    \includegraphics[width=0.09\textwidth]{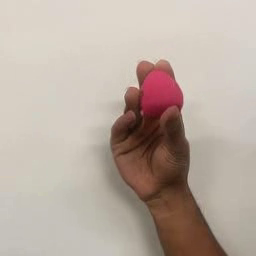} \\ [4.0pt] 
    && Reference& \multicolumn{4}{c}{Time} \vspace{.5em}\\
    &&& \multicolumn{4}{c}{
        \begin{tikzpicture}
            \draw[->, line width=0.3mm] (0,0) -- (6,0);
        \end{tikzpicture}} \\
    \end{tabular}
    \caption{Without explicit object supervision, \modelname synthesizes physically plausible object behaviors, including rigid motion (top: move cup) and non-rigid deformation (bottom: squish sponge), demonstrating an implicit understanding of object properties and dynamics.}
    \label{fig:hoi}
\end{figure}

\begin{figure}[!tp]
    \renewcommand{\tabcolsep}{0.0pt}
    \renewcommand{\arraystretch}{0.0}
    \centering \footnotesize
    \begin{tabular}{ccccccc}
    \rotatebox{90}{\hspace{0.5cm} \footnotesize\cite{controlnext}} & 
    \; &
    \includegraphics[width=0.09\textwidth]{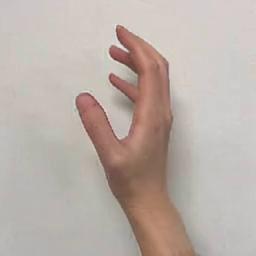} & 
    \includegraphics[width=0.09\textwidth]{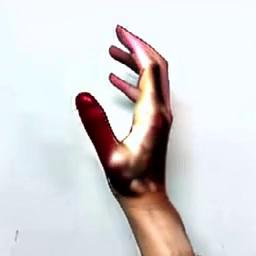} & 
    \includegraphics[width=0.09\textwidth]{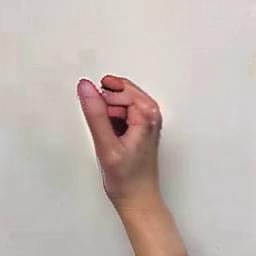} & 
    \includegraphics[width=0.09\textwidth]{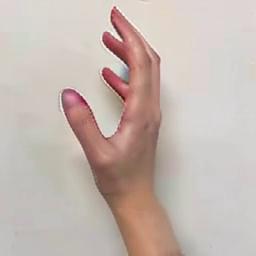} & 
    \includegraphics[width=0.09\textwidth]{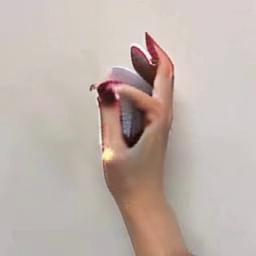} \\
    \rotatebox{90}{\hspace{0.5cm}\footnotesize \cite{animate_anyone}\;}&
    &
    \includegraphics[width=0.09\textwidth]{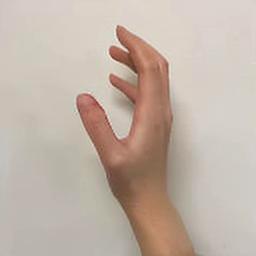} & 
    \includegraphics[width=0.09\textwidth]{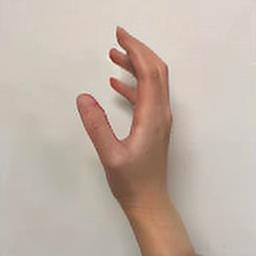} &
    \includegraphics[width=0.09\textwidth]{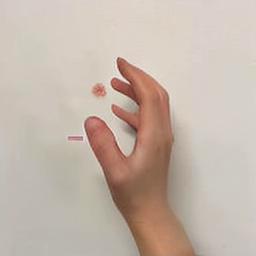} & 
    \includegraphics[width=0.09\textwidth]{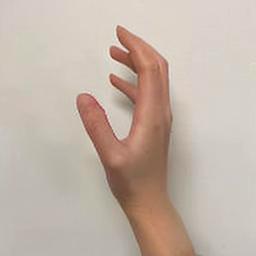} & 
    \includegraphics[width=0.09\textwidth]{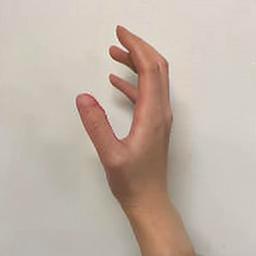} \\
    \rotatebox{90}{\hspace{0.5cm}\footnotesize Ours\;}&
    &
    \includegraphics[width=0.09\textwidth]{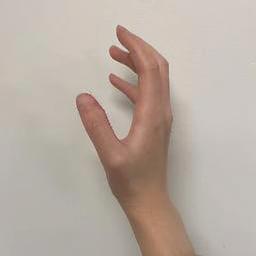} &
    \includegraphics[width=0.09\textwidth]{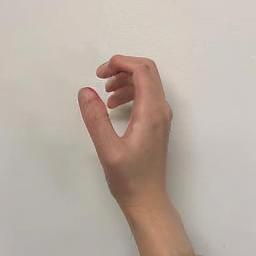} &
    \includegraphics[width=0.09\textwidth]{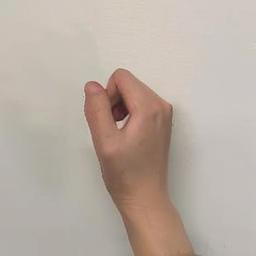} &
    \includegraphics[width=0.09\textwidth]{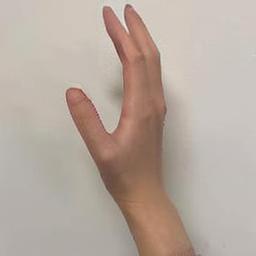} &
    \includegraphics[width=0.09\textwidth]{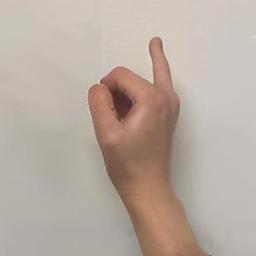} \\
    \rotatebox{90}{\hspace{0.7cm}\footnotesize GT\;}&
    &
    \includegraphics[width=0.09\textwidth]{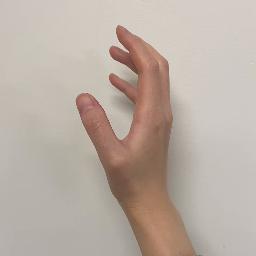} &
    \includegraphics[width=0.09\textwidth]{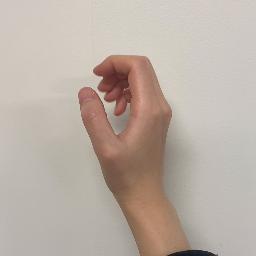} &
    \includegraphics[width=0.09\textwidth]{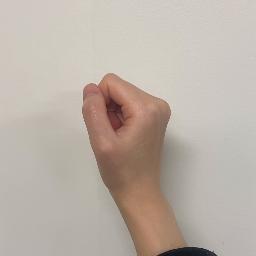} &
    \includegraphics[width=0.09\textwidth]{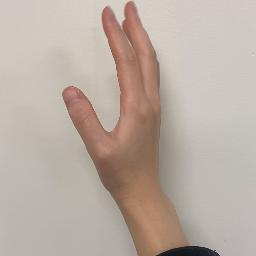} &
    \includegraphics[width=0.09\textwidth]{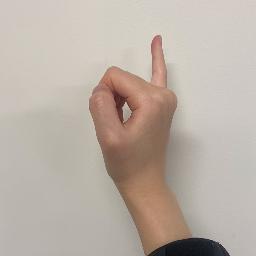} \\[3.0pt]
    && Reference& \multicolumn{4}{c}{Time} \vspace{.5em}\\
    &&& \multicolumn{4}{c}{
        \begin{tikzpicture}
            \draw[->, line width=0.3mm] (0,0) -- (6,0);
        \end{tikzpicture}} \\
    \end{tabular}
    \caption{\modelname achieves more natural and anatomically plausible hand motions without explicit video training.}
    \label{fig:video}
\end{figure}

\section{Applications}
The core gesture and domain transfer, and NVS capabilities of our \modelname enable it to be useful for downstream tasks -- we provide three examples below.

\paragraph{HandFixer.}
State-of-the-art text-to-image models consistently struggle with hand generation, often producing anatomically incorrect hands with malformed fingers.
While previous approaches~\cite{lu2023handrefiner,handcraft,realishuman} attempt to address this by incorporating 3D template meshes~\cite{mano}, their effectiveness is limited by the reliability of 3D estimation.
Moreover, existing methods like~\cite{lu2023handrefiner} struggle to maintain consistency in hand appearance, accessories, and handheld objects.
We demonstrate that \modelname can be directly adapted for fixing malformed hands without any task-specific training and offers more flexible user control through simple 2D keypoint and mask drawing interface. 
Our approach follows a similar strategy as Repaint~\cite{repaint} --
during the reverse denoising process, we sample $z_{\tau-1} = m\odot \bar{z}_{\tau-1} + (1-m)\odot z_{\tau-1}^0$, where $z_{\tau-1}^0$ from the known area of the original image via forward diffusion while $\bar{z}_{\tau-1}$ is sampled from \modelname given the previous step $z_{\tau}$.
By conditioning only on the specified keypoints without the reference condition, our model leverages its learned hand prior to generate anatomically correct hands. 
In \Cref{fig:handfixer} and supplemental, our approach shows outstanding generalization across diverse styles (digital art, paintings, sketches) and reliably preserves or inpaints any interacting objects. 
In contrast, task-specifically trained \cite{realishuman} often ignores contextual objects, and \cite{lu2023handrefiner} tends to hallucinate unwanted details and fails to maintain image consistency.

\paragraph{Motion-Controlled Video Synthesis.} 
Given a first frame and a sequence of 2D keypoints representing hand motion, our model enables video synthesis.
We observe that simply conditioning each frame generation on its immediate predecessor leads to artifact accumulation and quality degradation over time while using only the first frame as reference results in temporal inconsistencies. 
To address this, when generating frame $\mathcal{I}_t$, at each denoising step we randomly select the reference condition from a set of frames including the first frame and the last k generated frames $\{\mathcal{I}_1, \mathcal{I}_{t-1}, \mathcal{I}_{t-2}, ..., \mathcal{I}_{t-k}\}$. 
Note that only $\mathcal{I}_1$ is given, and others are generated frames.
This stochastic conditioning strategy is adopted by our novel view synthesis approach and provides both long-term consistency with the original scene through the first frame and short-term temporal coherence through recent frames, having better stability and quality than relying only on the immediate predecessor $\mathcal{I}_{t-1}$.
In \Cref{fig:video} and \Cref{tab:nvs_video}, we evaluate FoundHand's zero-shot video synthesis capabilities on 12 in-the-wild hand videos, casually captured with a mobile phone.
Compared to pose-conditioned video diffusion models of ControlNeXt~\cite{controlnext} and Animate Anyone~\cite{animate_anyone}, our model demonstrates superior temporal coherence and motion plausibility in both quantitative metrics and qualitative results, despite not being explicitly trained on video.

\paragraph{Hand-Object Interaction Video Synthesis}
In Fig.~\ref{fig:hoi} and the Supplement, notably, our model exhibits an emergent understanding of hand-object interactions without any explicit object supervision. 
Given only an initial frame of hand-object interaction (e.g., holding cups or sponges) and subsequent hand poses, the model synthesizes physically plausible object transformations including translation and non-rigid deformation. 
This suggests that FoundHand has implicitly learned to model hand-object interaction dynamics through its training data. 
On the other hand, CosHand~\cite{coshand} was trained specifically on pairs of before- and after- object deformation by hands.
However, our method robustly outperforms ~\cite{coshand}, while~\cite{coshand} occasionally shows totally random objects such as pink bowl or green ball that could have been included in their training dataset, indicating high generalization ability of FoundHand.

%% file: sec/6_conclusion.tex
\section{Conclusion}
\label{sec:conclusion}

Despite the significant advancements in generative models, generating hands remains a challenging task.
This is primarily due to the intricate finger articulation and the scarcity of large, unified hand datasets that can effectively capture this complexity.
In this paper, we identify 2D keypoint heatmaps as readily scalable representation and build FoundHand10M - a large unified hand dataset by embracing diversity from 12 existing datasets.
With that, we present FoundHand - a large-scale learning model that creates photo-realistic hand images.
Throughout experiments, FoundHand shows remarkable generalization ability to any hand images, versatility, and zero-shot emergent ability, which includes outperforming task-specific baselines by a large margin in 6 different hand-related tasks.

\paragraph{Limitations and Future Work.}
Due to the latent resolution of 32x32, \modelname sticks to 256x256 size of inputs and outputs. While we can easily crop or resize around hands, it would be worth trying image super-resolution techniques to support higher resolution hand generation. In the future, we envision \modelname to contribute to digital human hand avatars in XR applications and inverse problems like hand pose estimation. 



\paragraph{Acknowledgment.} Part of this work was done during Kefan (Arthur) Chen’s internship at Meta Reality Lab. 
This work was additionally supported by NSF CAREER grant \#2143576, NASA grant \#80NSSC23M0075, and an Amazon Cloud Credits Award.